\documentclass{amsart}
\usepackage[utf8]{inputenc}
\usepackage[foot]{amsaddr}
\usepackage{appendix}
\usepackage{mathtools}

\title[Optimising Stochastic Routing for Taxi Fleets with Model-enhanced RL]{Optimising Stochastic Routing for Taxi Fleets with Model Enhanced Reinforcement Learning}

\author{Shen Ren}
\address[Shen Ren, Liye Zhang, Zheng Qin]{Institute of High Performance Computing, A*STAR, Singapore}
\email{renshen@ihpc.a-star.edu.sg}
\author{Qianxiao Li}
\address[Qianxiao Li]{Department of Mathematics, National University of Singapore, Singapore}
\email{matlq@nus.edu.sg}

\author{Liye Zhang}
\author{Zheng Qin}

\email{zhangly@ihpc.a-star.edu.sg, qinz@ihpc.a-star.edu.sg}

\author{Bo Yang} \thanks{The last author is the corresponding author.}
\address[Bo Yang] {School of Physical and Mathematical Sciences, Nanyang Technological University, Singapore}
\email{yang.bo@ntu.edu.sg}

\usepackage{color}
\usepackage[numbers]{natbib}
\usepackage{graphicx}
\usepackage{tipa}
\usepackage{bm}
\usepackage{amsmath}
\usepackage{amssymb}
\usepackage{amsthm}
\usepackage{amsfonts}
\usepackage{float}
\usepackage{enumitem}
\usepackage{makecell}
\usepackage{multirow}
\usepackage{verbatim}
\usepackage{graphicx}
\usepackage{widetext}
\usepackage{flushend}
\usepackage{cuted}
\usepackage[space]{grffile}
\usepackage{ulem}

\begin{document}

\maketitle

\begin{abstract}
    The future of mobility-as-a-Service (Maas)
    should embrace an integrated system of ride-hailing, street-hailing and ride-sharing with optimised intelligent vehicle routing in response to a real-time, stochastic demand pattern. We aim to optimise routing policies for a large fleet of vehicles for street-hailing services, given a stochastic demand pattern in small to medium-sized road networks. A model-based dispatch algorithm, a high performance model-free reinforcement learning based algorithm and a novel hybrid algorithm combining the benefits of both the top-down approach and the model-free reinforcement learning have been proposed to route the \emph{vacant} vehicles. We design our reinforcement learning based routing algorithm using proximal policy optimisation and combined intrinsic and extrinsic rewards to strike a balance between exploration and exploitation. Using a large-scale agent-based microscopic simulation platform to evaluate our proposed algorithms, our model-free reinforcement learning and hybrid algorithm show excellent performance on both artificial road network and community-based Singapore road network with empirical demands, and our hybrid algorithm can significantly accelerate the model-free learner in the process of learning.
\end{abstract}

\section{Statement of Contribution/Potential Impact}

In this paper, we propose methodologies for optimising vacant vehicle routing in a decentralised way to meet stochastic spatio-temporal demand patterns. The research is motivated by the grand challenge of urban mobility to meet increasing demands for highly efficient yet high-quality transport services with minimal costs. The difficulty of the problem stems from the stochastic spatio-temporal distribution of the commuter demand, on a large road network with complex structures.
We design a set of methodologies for intelligent routing of taxi fleet with reinforcement learning primed with a top-down dispatch model, evaluated on both synthetic and empirical data for commuter demand patterns and road networks. The proposed methodologies can potentially benefit everyone in tackling the urban mobility challenges: shorter waiting time for commuters, higher profits for service providers, easy-to-implement and versatile solutions for government agencies, and less emissions to the environment. The methodologies presented in this paper are in time for real-world applications under current technology and in near future when wireless communication and autonomous vehicles enable new ways of taxi fleet operation. They could also have wider applications beyond taxi systems, for on-demand transportation and logistics services such as dynamic reallocation of shared resources and last-mile delivery.

\section{Introduction}
\label{sec:intro}

Diversified transportation services are needed in urban cities with high population density to meet the demands of commuters with high-quality services --- to transport from location A to B within designated time and with minimal costs. Traditional public transports including scheduled trains and buses serve large groups of commuters simultaneously, but follow fixed routes and pickup/drop-off locations on scheduled time, mostly in central areas where demand is high. Private vehicles on the other hand provide highly flexible and customised services for short to medium distance travel requests. But the utilisation rate for private vehicles is low, leading to high cost and increasing environmental concerns. The scalability for private vehicles is also highly constrained especially in metropolitan cities due to limited parking space and traffic congestion. 
 
In recent decades, there have been substantial behavioural changes in using mobility as a service in urban cities. These include shared vehicles, bridging buses and real-time, on-demand shuttles. These bridging services fill the gap between scheduled public transports and private vehicles, that can balance between service efficiency and service quality. The idea of demand responsive transit (DRT) is not new, but attracts great attentions in recent years for the purpose of efficient city planning and environmental concerns.
While the first DRT trial is implemented in Mansfield, Ohio, USA in 1970 \cite{oxley1980dial}, there has been a resurgence of DRT applications and trials recently throughout the world, especially in light of the new technological development including GPS, GIS solutions, tele-communications and smart devices. Examples including trials at Nijmegen-Arnhem region and Amsterdam of the Netherlands in 2016\cite{jittrapirom2019dutch} and 2017 \cite{coutinhoimpacts},  Inner East Sydney of Australia in 2018\cite{perera2019resurgence}, Sutton and Ealing of UK in 2019 \cite{ealing2019} and Singapore in 2018\cite{jin2019mobility}. Meanwhile, DRT applications have been especially popular in airport transportation \cite{reinhardt2013synchronized}, healthcare \cite{lim2017pickup}, and disruption handling \cite{jin2014enhancing}.

While DRT systems have been on trial for more than $50$ years, some significant challenges hindered a wide adoption of DRT systems in urban transportation services. Firstly, most of DRT systems operate under a Dial-A-Ride scheme, requiring commuters to phone in with deterministic pickup and drop-off requests ahead of time. Even with the pervasive use of smartphones, the requirement of the commuters to have advanced travel plans is in many cases a big constraint. Secondly, DRT systems are designed to complement the scheduled train and bus services. Thus they often operate in relatively lower demand areas. Without an efficient planning, scheduling and routing algorithm, it is difficult for DRT to provide high-quality services. Thirdly, DRT operating on large-scale complex networks requires a high-level central management, operation, booking and tracking system. This in turn pushes up the operation cost, rendering it economically infeasible for many potential commuters.

On the other hand, taxis as a form of on-demand services has been successfully serving urban cities for more than $100$ years. The taxi services can cater to commuters without the necessity of advanced booking or planning. Street-hailing is always a choice for people in need or people with unscheduled demands. The taxi system is also inherently distributive, since each taxi driver routes his/her vehicle independently to optimise profits based on personal experience. Thus the taxi service can achieve both lower operation costs and high flexibility, as compared to other DRT services. However, there is still much room for improvement, since in general the cruising time for taxi is still long compared to the hiring time \cite{hoque2012analysis}
and at the same time, commuters can face difficulties hailing a taxi especially during peak hours \cite{wang2018understanding}.
This is mainly because the demand and supply of the taxi services are not optimised in terms of their spatio-temporal distributions, due to individual taxi driver's lack of information on the real-time dynamics of the demand and other drivers' behaviours. Under-utilisation of the taxis is also a major cause of traffic congestion, with a lot of additional trips by vacant taxis looking for potential commuters.

The future of ``Mobility-as-a-Service'' (MaaS) should embrace a combination of DRT system and taxi services --- integrating ride-hailing, street-hailing and ride-sharing with highly-optimised intelligent vehicle routing to enhance service quality and to maintain cost-effectiveness. Emerging technologies including wireless communication and autonomous vehicles allow innovative ways of operating MaaS with novel algorithms that were hard to implement in a traditional manner. Individual vehicles can thus be highly flexible yet globally optimised to balance demand-and-supply with new technologies expected in the not too distant future.

In this paper, we present our first work towards next generation urban mobility with on-demand services. We focus on distributed vehicle routing algorithm of MaaS with stochastic spatio-temporal demand patterns. Our methodology combines both the top-down model construction, and the bottom up data-driven reinforcement learning (RL) approach, to develop a turn-by-turn routing algorithm for free agents (e.g. the vacant taxis) so as to efficiently seek potential customers to maximise jobs and to minimise waiting time of customers. The de-centralised nature of our method allows the supply of free agents to match the non-trivial stochastic demand over a complex network in a self-organised way, which can be easily scaled to large systems. The algorithm we develop can be applied to many transportation and logistics problems with stochastic requests such as dynamic shared bicycle re-allocation and last-mile delivery. Throughout this work, we will use the taxi system as the example for both the development of the methodology and for the agent based simulations. 

More specifically, we start with the construction of a centralised dispatch algorithm for vacant taxis, which can dispatch them to hot-spot locations in the road network where there is a higher probability of picking up potential commuters. From extensive microscopic agent based simulations, the dispatch algorithm is used to generate stochastic model-based turn-by-turn routing policies for each taxi. Both the dispatch algorithm and the model-based policy are used as benchmarks for model-free turn-by-turn manoeuvring policies from state-of-the-art RL algorithm. The latter leads to highly efficient distributed routing strategies for vacant taxis and at the same time requires minimal global information about the commuter demand, which is hard to obtain in realistic settings. A hybrid model-free RL with model-based initialisation algorithm is then proposed, that incorporates the benefits of both the model-free RL, and an optimised initial estimation from model-based policies. Numerical experiments were conducted with agent-based modelling on artificial lattice road networks and community-based Singapore road network with both synthetic and empirical demand patterns. Experiments showed the effectiveness of the proposed dispatch algorithm, model-based effective policies, model-free RL and hybrid algorithm. Among them, the proposed hybrid algorithm showed best performance in both artificial network and Singapore road network, and can significantly accelerate the model-free learner in the process of training.


In the following parts of this paper, we begin with related works reviewed in Sec.\ref{sec:related works}. The research problem is formulated in Sec.\ref{sec:problem} and the simulated environment and the Markov Decision Process are illustrated in Sec.\ref{sec:system}. Sec.\ref{sec:methodology} gives a thorough explanation on our proposed model-based methods, model-free RL and the hybrid approach. Sec.\ref{sec:experiment} shows a brief summary of our experimental design and results are presented in Sec.\ref{sec:results}. Finally, we discuss our results and concludes with future outlooks at Sec.\ref{sec:discussion}.

\section{Related Works}
\label{sec:related works}

Vehicle routing problem is the general problem of dispatching a fleet of vehicles to serve a given set of demands, first proposed in \cite{dantzig1959truck}. Taxi routing as a special case of vehicle routing problem, consists of assigning $m$ homogeneous vehicles and designing routes for $n$ given commuters with specified pickup and drop-off requests between origins and destinations. The street-hailing taxis and DRT routing problem have long been studied under Dial-a-Ride Problem (DARP) \cite{cordeau2007dial}. Depending on whether all the requests are known beforehand or dynamically revealed during the service period, DARPs can be divided into static and dynamic cases, in which the vehicle routes are either pre-designed or real-time adjusted to meet demands. Depending on whether the information received at the time of decision is certain without imperfect information or further changes, DARP can be classified into deterministic and stochastic cases \cite{ho2018survey}. The optimisation objectives of this problem can be either minimising costs subject to full demand satisfaction or maximising demand satisfactions subject to vehicle supplies \cite{cordeau2007dial}. 

Within the four categories, the dynamic stochastic multi-vehicle DARP, applied to ride-hailing and delivery services, is considered to be one of the most difficult to solve, where the demands are revealed during service period and subject to dynamic traffic conditions, service request changes and cancellations. Various algorithms have been proposed to optimise service quality of real-life dynamic DARP applications while meeting the constraint of vehicle capacity, maximum route duration and service time for small-sized instances, including heuristics and fuzzy-logic \cite{madsen1995heuristic, teodorovic2000fuzzy}, branch-and-bound \cite{colorni2001modeling}, integrated offline and online phases for new request insertion \cite{coslovich2006two}, adaptive large neighbourhood search \cite{gschwind2019adaptive}, hybrid tabu search and constraint programming \cite{berbeglia2012hybrid}. A prediction module is proposed to solve stochastic information in dynamic DARP, which is to use given imperfect information to predict future scenarios needed for decision making \cite{nunez2014multiobjective, munoz2015methodology}. Some variants of the dynamic DARP have also been investigated to understand the lower bound of competitiveness ratio, for example, Uber problem \cite{dehghani2017stochastic} and online k-taxi problem \cite{coester2018online}. In recent years, emerging technologies including electrical vehicles and autonomous vehicles have opened up new opportunities and also imposed new challenges to the traditional DARP \cite{narayanan2020shared}. Some prospective studies have presented the formulation and algorithms for electric autonomous DARP on small instances \cite{bongiovanni2019electric}, and for static autonomous DARP with ride-sharing variant and dynamic traffic \cite{liang2020automated}.

In practice, taxi services includes both ride-hailing and street-hailing. The vacant taxis route to maximise demand satisfaction with pickup drop-off service requests revealed during service period either via ride-hailing (the taxi central management will assign taxis for booking requests using various taxi assignment algorithms or heuristics) or via street-hailing instantaneously. The challenge of street-hailing is that the demands are not explicitly phone-in by commuters but revealed only until a vacant taxi happens to pass by the commuters' origins before the commuters change their minds. The existing solutions mainly consists of two steps: first, to construct a model for demand prediction with a quantification of uncertainty; then, to formulate the routing problem with prediction uncertainty as an optimisation problem and solve by exact methods, heuristics and meta-heuristics \cite{miller2017demand, feng2017we, yu2019markov}. A very recent study presented centralised and decentralised dispatch algorithms for autonomous taxis that can serve both reserved requests and immediate requests after departure using a network flow model \cite{duan2020centralized}. In comparison with this work, our model-free RL approach does not require pre-knowledge of the demand patterns for immediate requests prediction, and our model-based approach does not require exact requests to be revealed to taxis. There are also a few studies apply RL and formulates the vacant taxi routing as a Markov Decision Process to plan for long-term rewards \cite{han2016routing, verma2017augmenting, yu2019markov}, including our previous work \cite{yang2018turn}. Compared with the existing studies and our previous work, we have significantly enhanced our routing performance by algorithm design and the hybrid approach allows for model-free RL fine-tuning upon existing methods with better final performance and significant higher sample-efficiency.

Though RL has not been widely applied in vacant taxi routing, it is widely considered as a strong AI paradigm with successful applications in games \cite{mnih2013playing, silver2017mastering}, robotics, navigation \cite{levine2013guided, zhu2017target}, natural language processing, computer vision, healthcare, smart grid, train rescheduling \cite{vsemrov2016reinforcement} and a lot of others. RL consists of a group of algorithms with their own advantages and trade-offs for different problems. A RL problem can be formulated with solution methods model-based or model-free, value-based or policy-based, on-policy or off-policy, with function approximation or not, and with sample backups or full backups. In general, model-based RL method learns value function to model the environment that the agents interact with, while model-free RL deals with unknown dynamical systems. Continuous attempts are proposed to combine the advantages of model-free and model-base RL approaches, including model predictive control \cite{camacho2013model} and guided policy search \cite{levine2013guided}. Various advanced approaches have also been proposed for reinforcement learning with function approximators, including some successful applied algorithms like deep Q-learning \cite{mnih2015human}, policy gradient method \cite{mnih2016asynchronous}, trust region policy gradient method \cite{schulman2015trust}, and proximal policy optimisation \cite{schulman2017proximal}. The fundamental trade-off of RL is exploration-exploitation. Designing better exploration strategy working for different RL schemes and problems is also an active area at the moment \cite{osband2016deep, burda2018exploration}.

\section{Problem Formulation}
\label{sec:problem}

In our framework, an agent-based model simulates a fleet of taxis roaming on the road network based on a stochastic routing policy, picking up and delivering commuters generated in the road network from their origins to their respective destinations. The positions, behaviours and status of the taxis (vacant or occupied) are recorded and used as inputs for a separate RL algorithm, which is learned to optimise the routing policy in an iterative manner. 
The challenge in tackling this complex system with a large number of agents is to have a mathematical formulation that satisfies several requirements. Firstly, the formulation has to be flexible enough to easily incorporate realistic details of the transportation system. Secondly, it should allow highly efficient numerical simulations for reasonably large road network containing thousands of nodes. There are also generally up to thousands of taxis roaming in the network with sophisticated behaviours, delivering up to tens of thousands of commuters from their origins to destinations over each simulation period. Thirdly, the formulation needs to be fully integrated to RL algorithms, where the behaviours of the taxis as well as the ``rewards" of their actions (to be defined more precisely later on) can be easily collected and analysed by the modern machine learning algorithm.

To that end, we abstract the road network with a connected directed graph denoted as $G\left(\mathcal N,\mathcal E\right)$, where $\mathcal N$ represents the collection of nodes of the graph, and $\mathcal E$ is the collection of directed weighted edges. The nodes correspond to all points-of-interest of a region (i.e. a city) where commuters can board or alight taxis. The edges are represented by an adjacency matrix $\mathcal A$ characterising the travel time on each road segment with respect to traffic conditions. For nodes pair $i,j$, the node $j$ is directly connected from node $i$ if and only if the taxi can drive from $i$ to $j$ without passing through any other nodes in $G$. 
In our definition of the adjacency matrix $\mathcal A_{ij}$, if $j$ is not directly connected from $i$, $\mathcal A_{ij}=0$; otherwise for $i\neq j$, $\mathcal A_{ij}$ is the \emph{reciprocal} of the travel time from $i$ to $j$. When $i=j$, $\mathcal A_{ii}=1$.

Two arrays are used to characterise commuters' behaviours: $g_i$ gives the probability of a commuter appearing and waiting at node $i$ for each time step; $M_{ij}$ gives the probability of a commuter with origin at node $i$ is going to the destination at node $j$. The constraints for these two arrays are given in Eq.(\ref{mm}) and Eq.(\ref{gg}). In principle, $\mathcal A_{ij}, g_i, M_{ij}$ can all depend on time. In this paper we focus on the stationary case, and we will leave the time-dependent cases for the future works.

The routing algorithm we develop is entirely encapsulated in a stochastic policy matrix $\mathcal P_{ij}$ for vacant taxis, which is also the decision variable of the optimisation. The stochastic routing policy defines the the probabilities of moving to a directly connected node $j$ when a vacant taxi is located at $i$. If $\mathcal A_{ij}=0, \mathcal P_{ij}=0$. The $\mathcal P_{ii}$ gives the probability of a vacant taxi staying at node $i$ for one time step, which can be non-zero. For hired taxis, the routing follows shortest path on $\mathcal A_{ij}^{-1}$. A simple initial policy is the random policy, defined as $\mathcal P_{ij}=1/n_i$ if $\mathcal P_{ij}\neq 0$. Here $n_i$ is the number of neighbouring nodes for each $i$ where $\mathcal A_{ij} \neq 0$, including $j=i$.

Let $n$ be the total number of commuters generated at each time step, the inputs for the large scale agent-based simulations are $\mathcal A_{ij},g_i,M_{ij},\mathcal P_{ij}$ subjecting to the following constraints:
\begin{align}
&\sum_{j \in \mathcal A_{ij} \neq 0} \mathcal P_{ij} = 1, & \forall i \in \mathcal N \\
& \sum_{j \in \mathcal N} M_{ij} = 1, & \forall i \in \mathcal N \label{mm}\\
&\sum g_{i} = n_c , & \forall i \in \mathcal N \label{gg}\\
& \mathcal P_{ij} \in \{0, 1\}, M_{ij} \in \{0, 1\}, g_i \in \{0, 1\} &\forall i,j
\end{align} 

There are two types of agents in the simulation: the taxis and the commuters. The behaviour of the commuters is rather simple. At every time step, we scan over every node in the network. There is a probability $g_i$ of increasing the queuing commuter at node $i$ by 1. We can thus denote $c_i$ as the number of commuters waiting at node $i$. The commuters do not move around the network, and $c_i$ decreases by 1 when a vacant taxi reaches node $i$. Thus at time step $t$, the dynamical behaviour of $c_i$ is given by:
\begin{eqnarray}
\qquad c_i\left(t\right)=\left\{
\begin{array}{ll}
      \text{max}\left(c_i\left(t-1\right)+1-N_{\text{vacant}}\left(i,t\right),0\right) & \text{with probability }g_i \\
      \text{max}\left(c_i\left(t-1\right)-N_{\text{vacant}}\left(i,t\right),0\right) & \text{with probability }1-g_i \\
\end{array} 
\right.
\end{eqnarray}
where $N_{\text{vacant}}\left(i,t\right)$ is the number of vacant taxis at node $i$ at time $t$. For total simulation time $H$, the total waiting time of the commuters can be calculated as:
\begin{eqnarray}
T_{\text{waiting}}=\sum_{t=0}^H\sum_{i=1}^{\mathcal N}c_i\left(t\right)
\end{eqnarray}

Let $n_c$ represents number of commuters generated at each time step, the average waiting time of all commuters is given by:
\begin{eqnarray}
\bar{T}_{\text{waiting}}=\frac{T_{\text{waiting}}}{H n_c}
\label{eq:avg_waiting}
\end{eqnarray}

The dynamics for the taxis are more complicated. We denote the location of the $l^{\text{th}}$ taxi by a time-dependent array $\left(k^l_1,k^l_2,d^l,s^l\right)(t)$, when the taxi is in between two connected nodes $k^l_1,k^l_2$ and moving towards $k^l_2$, with time distance $d^l$ from $k^l_1$ (so the taxi is $d^l$ time steps away from $k_1^l$). Thus $d^l\le \mathcal A_{k^l_1k^l_2}^{-1}$. If $d^l=0$, the taxi is right at the node $k^l_1$. We use $s^l$ to indicate the status of the taxi, which in our simulation is either vacant ($s^l=v$) or occupied ($s^l=o$). We clearly have $\left(k^l_1,k^l_2,A_{k_1k_2}^{-1},s^l\right)(t)=\left(k^l_2,k^l_3,0,s^l\right)(t)$. The formal dynamical equations are given as follows:
\begin{eqnarray}
&&d^l\left(t\right)=\left\{
\begin{array}{ll}
      d^l\left(t-1\right)+1 & d^l\left(t-1\right)<\mathcal A^{-1}_{k_1^lk_2^l} \\
      0 & d^l\left(t-1\right)=\mathcal A^{-1}_{k_1^lk_2^l} \\
\end{array} 
\right.\\
&&\left(k_2^l,k_3^l,0,v\right)(t)= \left(k_1^l,k_2^l,0,v\right)(t-1)\qquad\text{if }c_{k_2^l}=0\\
&&\left(k_2^l,\bar k_3^l,0,o\right)(t)= \left(k_1^l,k_2^l,0,e\right)(t-1)\qquad\text{if }c_{k_2^l}>0
\end{eqnarray}
 If $s^l=v$ and no commuter is waiting at the node, $k^l_3$ is chosen with probability $\mathcal P_{k^l_2k^l_3}$; if $s^l=o$, then $\bar k^l_3$ is chosen from the shortest path to the destination of the commuter in the taxi. We can now define a reward function $R$ for the taxis based on whether or not there is a commuter in the taxi. A simple example we use in the paper is to give a unit reward to every occupied taxi for each time step. The reward function is further detailed in Sec.\ref{sec: MF RL}.

Now we can treat the task formally as an optimisation problem. Given a road network $\mathcal A_{ij}$, a stochastic taxi demand $g_i$ and $M_{ij}$, and a fixed supply of $N$ taxis, we would like to find the optimal taxi stochastic routing policy $\mathcal P_{ij}$ as decision variables that maximise or minimise certain objective functions that we can freely define. Different objective functions can be defined based on prioritising different parties involved (e.g. the passengers, the taxi drivers, and the policy-makers). In this paper, we choose to maximise total reward defined in Eq.(\ref{eq:reward}). The summarised notations are shown in Table.~\ref{t1}.
\begin{table}
\centering
\begin{tabular}{ |c|c|c| } 
 \hline
 Network & Commuter & Policy  \\ 
 \hline
 \multirow{2}{8em}{\centering Adjacency matrix $\mathcal A_{ij}$} & Origin probability distribution $g_i$ & \multirow{2}{6em}{\centering Policy matrix $\mathcal P_{ij}$}\\ 
 & Destination probability distribution $M_{ij}$ & \\ 
 \hline
\end{tabular}
\caption{Notations of the Formulated Problem}
\label{t1}
\end{table}



\section{System Overview}
\label{sec:system}

\subsection{Agent-based Simulation}
\label{sec:simulation}

We have designed and implemented a large scale microscopic agent-based simulation platform \cite{yang2020phase} for evaluation of the performance of $\mathcal P_{ij}$, using different $\mathcal A_{ij},g_i,M_{ij}$ as inputs. A collection of instances containing state information on the positions and actions of vacant taxis, together with relevant extrinsic rewards needed for reinforcement learning is generated from the simulation after running the simulation for finite time horizon $H$ (measured in time steps). While different physical time unit can be used to correspond to each time step, in this study, each time step is taken to be one second for generality. The platform simulates the dynamics of two types of agents: the commuters and the taxis, and produces outputs characterising the performance of stochastic policy $\mathcal P_{ij}$ from different perspectives using various performance metrics. We focus on average extrinsic reward and average waiting time of commuters in this paper, further explained in Sec.\ref{sec:experiment}.

The behaviours of commuters are determined by probability distributions $g_i$ and $M_{ij}$, and the initial positions of all taxis are assigned to random nodes at the beginning of each epoch of the simulation. Vacant taxis will only pick up a commuter when the location of a vacant taxi and the origin of a waiting commuter coincides. The movements of the occupied taxis are computed efficiently by shortest path algorithm using our customised Contraction Hierarchy algorithm \cite{yang2020phase}, while the movements of the vacant taxis will follow the stochastic routing policy $\mathcal P_{ij}$. Both real-time booking and road-side hailing with three potential status for the taxis: vacant, booked, and occupied, can be implemented. In this paper, we focus on street-hailing instead of real-time booking so taxis will not take bookings. But in scenarios where some of the taxis take bookings, the methods proposed in this paper can still be applied to route those vacant taxis available for both booking and street-hailing by potential commuters. 



\subsection{The Markov Decision Process} 

\begin{figure}[tbh]
\centering
\includegraphics[width=9cm]{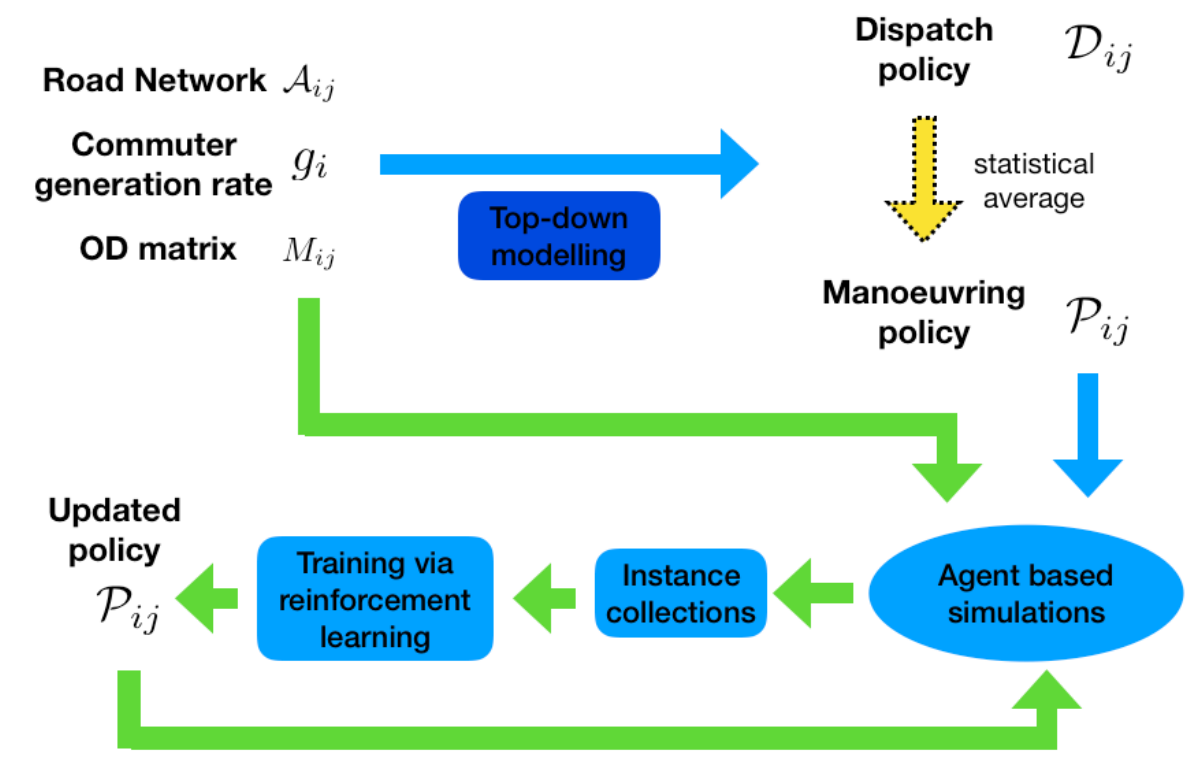}
\caption{An illustration of framework for optimising vacant taxi routing using model-based and model-free approaches.}
\label{fig:framework}
\end{figure}

The simulation platform can be used to evaluate policies $\mathcal P_{ij}$ from top-down construction using model-based methods, or from bottom-up learning using reinforcement learning. An illustration of our proposed framework for optimising taxi routing policies is shown in Figure \ref{fig:framework}.

The sequential decision makings of the taxis in above framework can be modelled as a Multi-agent Markov Decision Process (MMDP) involving a collective of taxis as agents, each with their own set of actions, operating in a common environment with the same shared goal (utility function) to achieve a fully cooperative system \cite{boutilier1996planning}. 
The challenge of a MMDP modelling of this problem (formulated in Sec.\ref{sec:problem}) lies in the set of actions to be represented and the policies to be optimised, as each vacant taxi will perform their own policies and choose actions at every location in the road network accordingly. While all agents are improving their own policies concurrently, the environment faced by each agent may become non-stationary.
And the joint action space increases exponentially with the number of agents \cite{zhang2019multi}, which may cause scalability issues in our problem when the number of taxis in systems is large.

To deal with the above challenges and to work for realistic taxi system with a large taxi fleet, we apply a "mean-field" approximation and model the given problem as a single-agent Markov Decision Process (MDP), assuming that each taxi will follow the same stochastic policy to be learned.
Instead of each taxi as an agent, a learner to optimise routing policies $\mathcal P_{ij}$ is formulated as the \textit{agent} and the simulation that the learner interacts with is the \textit{environment}. The resulting stochastic routing policies $\mathcal P_{ij}$ is used to determine the routing decisions for each vacant taxi in system.

Mathematically, the modelled finite MDP is specified by $(\mathcal S, \mathcal A, P, r, \gamma, \rho)$, where: $\mathcal S$ is a state space of the environment, $\mathcal A$ is an action space, $P$ is a transition model $P(s^\prime|s, a)$ as the probability of transition into state $s^\prime$ from $s$ by taking action $a$, $r$ is a reward function where $r(s, a)$ is the immediate reward in taking action $a$ in state $s$, $\gamma$ is a discount factor $\gamma \in [0,1]$, $\rho$ is a starting state distribution over $S$. In this problem, the state $\mathcal S$ is given as the state of all taxis in system. The action $\mathcal A$ are decisions made for vehicles at each node on which adjacent node to move to. As a consequence of the actions, the \textit{agent} receives reward $r=r(s_t, a_t), r \in \mathbb{R}$ characterising the optimisation objective defined in Sec.\ref{sec:problem}, and transit to the next state $s^\prime$ according to transition model $P$. The construction of the reward function is further explained in Sec.\ref{sec: MF RL}.

Let $\pi: \mathcal S \rightarrow \Delta (\mathcal A)$ be a stochastic policy specifies the decision-making strategy in choosing the probability of actions for each taxi in system based on current state, where $\Delta (\mathcal A)$ is the probability simplex over action space $\mathcal A$, and $a_t \sim \pi(\cdot|s_t)$. $\pi$ is represented as the policy matrix $\mathcal P_{ij}$ in this problem. By following $\pi$, a distribution of a state-action sequence $\tau = (s_0, a_0, \cdots , a_{H-1}, s_{H})$ is denoted as Eq.(\ref{eq: sadistribution}), starting from state $s_0$ drawing from starting state distribution $\rho$, and for all subsequent $i, i\in[0, H)$ where $a_i \sim \pi(\cdot|s_i)$ and $s_{i+1} \sim P(\cdot|s_i, a_i)$.

\begin{equation}
    D^{\pi}(\tau) \coloneqq \prod_{i=0}^{H-1} \pi(s_i, a_i) P(s_i, a_i, s_{i+1})
    \label{eq: sadistribution}
\end{equation}

Given a discount factor $0\leqslant \gamma \leqslant 1$, the value function for a policy $P_{ij}$ is defined as discounted sum of future rewards from starting state $s$ over a finite horizon $H$, i.e.

\begin{equation}
    V^{\pi} (s) \coloneqq \mathbb{E} \left[\sum_{t=0}^{H-1} \gamma^{t-1} r(s_t, a_t)|\pi, s_0=s \right]
\end{equation}

The expected value of initial state is further defined as Eq.(\ref{eq:valuefirststate}).

\begin{equation}
    V^{\pi}(\rho) \coloneqq \mathbb{E}_{s_0\sim \rho}[V^{\pi}(s_0)]
    \label{eq:valuefirststate}
\end{equation}

For a given policy $\pi$, the state-action value function (Q-value function) $Q^\pi$ and the advantage function $A^\pi$ are defined as Eq.(\ref{eq:qfunction}) and (\ref{eq:advantage}).

\begin{equation}
    Q^{\pi}(s,a) \coloneqq \mathbb{E}\left[\sum_{t=0}^{H-1} \gamma^{t-1} r(s_t, a_t)|\pi, s_0=s, a_0=a \right]
    \label{eq:qfunction}
\end{equation}

\begin{equation}
    A^{\pi}(s,a) \coloneqq Q^{\pi}(s,a) - V^{\pi}(s)
    \label{eq:advantage}
\end{equation}

The objective of the learning agent is to use ascent methods for the optimisation problem given as Eq.(\ref{eq:rlobjective}), which is to maximise expected value from the initial state over all policies.

\begin{equation}
    \max_{\theta \in \Theta} V^{\pi_\theta}(\rho)
    \label{eq:rlobjective}
\end{equation}

where $\pi$ in this study is parameterised by linear function approximator as

\begin{equation}
    \pi_\theta(a|s) = \theta_{s,a}
\end{equation}

where $\theta$ is a row stochastic matrix, $\theta_{s,a} \geqslant 0$ and $\sum_{a\in \mathcal A} \theta_{s,a} = 1$ for $s\in \mathcal S$ and $a \in \mathcal A$.

It has been proved in \cite{agarwal2019optimality} that $V^{\pi_\theta}(s)$ is non-concave in $\theta$ for the above parameterisation, so the standard solvers for convex optimisation problems are not applicable for this target problem. Throughout this paper, we will use RL approaches to optimise the routing policies based on policy gradient theorem \cite{sutton2000policy} as follows.

Let $R(\tau)$ be the total reward of state-action sequence $\tau$ on following $\pi$ starting from $s_0$, where in this problem, the reward $R(\tau)$ is given by

\begin{equation}
    R(\tau) \coloneqq R(s_0, a_0)
\end{equation}

where $R(s_0, a_0)$ is defined by Eq. (\ref{eq:reward}) in Sec.\ref{sec: MF RL}.

Then,

\begin{equation}
    \nabla_\theta V^{\pi_\theta}(\rho) = \mathbb{E}_\tau \left[ r(\tau)\nabla_\theta \log(D^{\pi_{\theta}}(\tau))|\rho\right]
    = \mathbb{E}_\tau \left[ r(\tau) \sum_{t=0}^{H-1} \nabla_\theta \log(\pi_{\theta}(s_t, a_t))|\rho\right]
    \label{eq:policygradient}
\end{equation}


Though the problem of this paper is formally framed as a MDP, the RL method proposed in this paper is based on Monte-Carlo method with undiscounted setting where $\gamma \rightarrow 1$, which does not require the system to be Markovian. The details of the model-free reinforcement learning will be further explained in Sec.\ref{sec:methodology}.

\section{Methodology}
\label{sec:methodology}
We present our proposed model-based approaches, model-free RL and hybrid learner in this section, based on the design shown in Figure \ref{fig:framework}. The performance of RL depends on 1) meaningful state space, action space and reward functions defined, 2) good designed algorithms to learn from data, and 3) efficient data collection by exploring the environment in a self-organised way from the above formulation. To achieve that, we first present our model-based routing policy constructed from top-down approach, which requires no sample data. We then demonstrate our design of RL algorithm and the construction of reward function. At the end, a simple and effective hybrid approach is described to initialise model-free RL with model-based effective policy.

\subsection{Model-based Routing Policy}
\label{sec: MB routing}



For this target problem, instead of learning a model of the dynamic function of environment (like in most model-based RL), we can construct routing policy from transportation theory using the MDP. In order to find the optimal $\mathcal P_{ij}$ which allows us to route a vacant taxi at any location in a given road network, the intuition behind our proposed model-based dispatch algorithm is to direct the vacant taxis to a set of ``hotspots" in the road network, which are identified either statically or dynamically. These hotspots are locations where the probability of finding a waiting commuter is high, so that vacant taxis are dispatched to these hotspots, which can be formally described by a dispatch matrix $\mathcal D_{ij}$: given a vacant taxi at node $i$, $\mathcal D_{ij}$ gives the probability of the taxi being dispatched to node $j$ (which potentially can be quite far away). The dispatch probability is also inversely proportional to the distance between $i$ and $j$. There will also be some routing choice from node $i$ to node $j$ (the simplest being the shortest path between the two nodes). Once $\mathcal D_{ij}$ and the routing choice are specified, the turn-by-turn routing of $\mathcal P_{ij}$ as model-based routing policy can also be uniquely determined via microscopic simulations and sampling.
Throughout this paper we assume shortest path routing between two nodes for $\mathcal D_{ij}$, as the specific choice of routing algorithm is not essential in this particular context. 

\subsubsection{Dispatch Algorithm for Hotspots}
One of the common approaches for vacant taxi manoeuvring is a good algorithm for dispatching them to certain locations on the map where higher probabilities of commuter generation are expected (the hotspots). This is also implicitly the strategy of experienced taxi drivers. Following the problem statements and definitions in Sec.~\ref{sec:problem}, we first propose models of efficient dispatch algorithm, using $\mathcal A_{ij},g_i,M_{ij}$ as inputs, to obtain an efficient dispatch matrix $\mathcal D_{ij}$. Physically, for a vacant taxi appearing at node $i$, $\mathcal D_{ij}$ gives the probability for this vacant taxi to be dispatched to a (not necessarily adjacent) node $j$, with the constraint $\sum_j\mathcal D_{ij}=1$. We also allow $\mathcal D_{ii}\neq 0$, implying the vacant taxi stay put at its location waiting for commuters to appear. If node $j$ is far away from node $i$, the vacant taxi will be dispatched to node $j$ via the shortest path obtained from the network adjacency matrix $\mathcal A_{ij}^{-1}$. Without loss of generality, here we only make $\mathcal D_{ij}$ dependent on $g_i$ and $\mathcal A_{ij}$, and not explicitly dependent on $M_{ij}$.

Let $d_{ij}$ be the shortest distance in terms of travel time between node $i$ and $j$, as computed from the adjacency matrix $\mathcal A_{ij}^{-1}$. For $i=j$ we set $d_{ii}=1$. A rather general dispatch model is given as follows:
\begin{eqnarray}\label{dispatchmodel}
\mathcal D_{ij}=f\left(d_{ij}^{-1},g_j\right)
\end{eqnarray}
Intuitively, $\mathcal D_{ij}$ should increase monotonically with $g_j$, while decreasing monotonically with $d_{ij}$. This is because if all other conditions are identical, we should have a higher probability of dispatching vacant taxis to nodes with higher rate of commuter generation, the so-called hotspots. On the other hand, for two nodes with the same rate of commuter generation, there should be a reduced probability to dispatch the vacant taxi to a node that is further away. Optimisation of Eq.(\ref{dispatchmodel}) can be done by Taylor expanding $f$ with respect to $d_{ij}^{-1}$ and $g_j$, and tuning the expansion coefficients at different orders to maximise the utility functions computed from agent based simulations. Here we use a rather simple model with the lowest order expansion as follows:
\begin{eqnarray}\label{dm1}
\mathcal D_{ij}=\lambda_i \frac{g_j}{d_{ij}}
\end{eqnarray}
where $\lambda_i=\left(\sum_j\mathcal D_{ij}\right)^{-1}$ is the normalisation factor. The performance of this model has been satisfactory based on extensive agent-based numerical simulations shown in later section Sec.\ref{sec:mb result}.

\subsubsection{Effective Routing Policy from Dispatch Model}

One major difference between $\mathcal D_{ij}$ and $\mathcal P_{ij}$ for vacant taxi manoeuvring is that for the former, nodes $i$ and $j$ can be quite far away. On the other hand, $\mathcal P_{ij}$ is only non-zero when nodes $i$ and $j$ are neighbours. The degrees of freedom for $\mathcal D_{ij}$ is thus much larger and scale as the square of the network size. This makes optimising $\mathcal D_{ij}$ a much more difficult problem, as compared to optimising $\mathcal P_{ij}$ (whose degrees of freedom scale linearly with network size). On the other hand, when the vacant taxi is dispatched from node $i$ to $j$, the path between nodes $i$ and $j$ is deterministic (and we choose the shortest path in this work). Effectively, given a specific $\mathcal A_{ij},g_i,M_{ij},\mathcal D_{ij}$ there is still a statistical distribution for the choice of neighbouring nodes from any node $i$, if we look at a large number of vacant taxis arriving at node i (either as the destination of the dispatch, or as an intermediate node enroute to the dispatch destination), and how these vacant taxis choose the next node. We can thus generate an effective $\mathcal P_{ij}$ via agent based simulations, and this particular $\mathcal P_{ij}$ is unique and well-defined in the limit of long time simulations.

Formally, for a simulation time with a total number of discrete time-steps $H$, we denote $n_i^H$ as the number of times when an vacant taxi lands on node $i$, and we do not distinguish between different taxis. Let $p_{ij}^H$ be the number of times the vacant taxi at node $i$ moves towards one of the neighbouring node $j$. The effective turn-by-turn routing policy is defined as follows:
\begin{eqnarray}
\bar{\mathcal P}_{ij}=\lim_{H\rightarrow\infty}\frac{1}{n_i^T}p_{ij}^H
\end{eqnarray}
In this way, the effective policy can be unambiguously defined for almost any manoeuvring strategies. If the routing strategy for vacant taxis is a turn-by-turn policy $\mathcal P_{ij}$ to start with, then obviously we will have $\bar{\mathcal P}_{ij}=\mathcal P_{ij}$. If the routing strategy is from a dispatch policy $\mathcal D_{ij}$, there is a unique one-to-one mapping from $\mathcal D_{ij}$ to $\bar{\mathcal P}_{ij}$, given all other conditions are fixed. We can thus evaluate and compare the performances of $\mathcal D_{ij}$ and $\bar{\mathcal P}_{ij}$ in a meaningful manner. More importantly, $\bar{\mathcal P}_{ij}$ can be used as an optimised initial state for reinforcement learning, as we will explain in the rest of this paper.

\subsection{Model-free Reinforcement Learning with Model-based Initialisation}

Top-down approaches in constructing $\mathcal P_{ij}$ can be highly efficient using no data samples on the states $s_t$, given a specific $\mathcal A_{ij},g_i,M_{ij}$, but they generally do not yield optimal turn-by-turn maneuvring policies. Model-free RL, on the other hand, can optimise the policies variationally, starting with any specific initial policy using policy gradient based methods. To combine the benefits of model-based approach and model-free RL, we propose to train a model-free RL from either random policy or the effective policy from dispatch algorithm as our initialisation policy. Thus, we can use RL to obtain high performance routing policies either initialised from a random policy without the information on $\mathcal A_{ij},g_i,M_{ij}$, or from model-based policies constructed from top-down approaches. 

\subsubsection{Model-free Reinforcement Learning}
\label{sec: MF RL}

Our model-free RL is initialised with some specific policies, which are then updated by Actor-Critic Method with linear function approximators and on-policy policy gradient \cite{sutton2018reinforcement}. To improve the performance and sample efficiency, Proximal Policy Optimisation (PPO) \cite{schulman2017proximal} is used to optimise a ``surrogate'' objective function using stochastic gradient ascent to update policy $\mathcal P_{ij}$ based on data sampled from $\mathcal P_{ij}$. An exploration bonus is further introduced to encourage the learning \textit{agent} to visit novel states even when the environmental extrinsic reward is sparse. 

\textbf{Policy Optimisation:}
More concretely, we employ Advantage Actor-Critic (A2C) \cite{mnih2016asynchronous} combined with PPO as a RL framework which serves as a well-performed baseline for a lot of discrete and continuous benchmarks, for example Atari, 3D humanoid, and MuJoCo Physics engine. In general, advantage actor-critic has two networks named as \emph{actor} and \emph{critic} accordingly. The actor is to learn the optimised policy using a policy network, as defined in Eq.(\ref{eq:policygradient}), and the critic is to evaluate the policy created by the actor and suggest the adjustment using a value network with calculated advantage, as defined in Eq.(\ref{eq:qfunction}) and (\ref{eq:advantage}. PPO is to used to reduce the variance of vanilla policy gradient by reducing reinforcement learning to a numerical optimisation problem on the policy.

In essence, the policy matrix $\mathcal P_{ij}(\theta)$ is parameterised by actor parameter vector $\theta$, where $\theta$ are trainable weights to maximise future rewards. An ``surrogate'' objective is maximised subject to the constraint that the distribution of the policy before update $\mathcal P_{ij}(\theta_{old})$ is close to the distribution of update policy $\mathcal P_{ij}(\theta)$. Hence, the objective can be modified to penalise changes to the policy that moves too far from the $\mathcal P_{ij}(\theta_{old})$. Formally, the objective to be optimised is shown as:

\begin{equation}
L^{\text{clip}}(\theta) = \mathbb{\hat E}_t[\min (r_t(\theta)\hat A_t, \text{clip}(r_t(\theta)), 1-\epsilon, 1+\epsilon)\hat A_t]
\label{eq:ppo}    
\end{equation}

where $\epsilon$ is a small hyperparameter to be tuned, $r_t(\theta)$ is the probability ratio $\frac{\mathcal P_{ij}(\theta)(a_t|s_t)}{\mathcal P_{ij}(\theta_{\text{old}})(a_t|s_t)}$. $\hat A_t$ is the estimated advantage at time $t$ from the critic, which captures how a particular policy is better than the others at a given state. The expectation in equation \ref{eq:ppo} is the empirical average of samples from the distribution of ${\mathcal P_{ij}(\theta_{\text{old}})(a_t|s_t)}$, using importance sampling to estimate expectation of $\mathcal P_{ij}(\theta)(a_t|s_t)$. 

\begin{figure}[tbh]
\centering
\includegraphics[width=9cm]{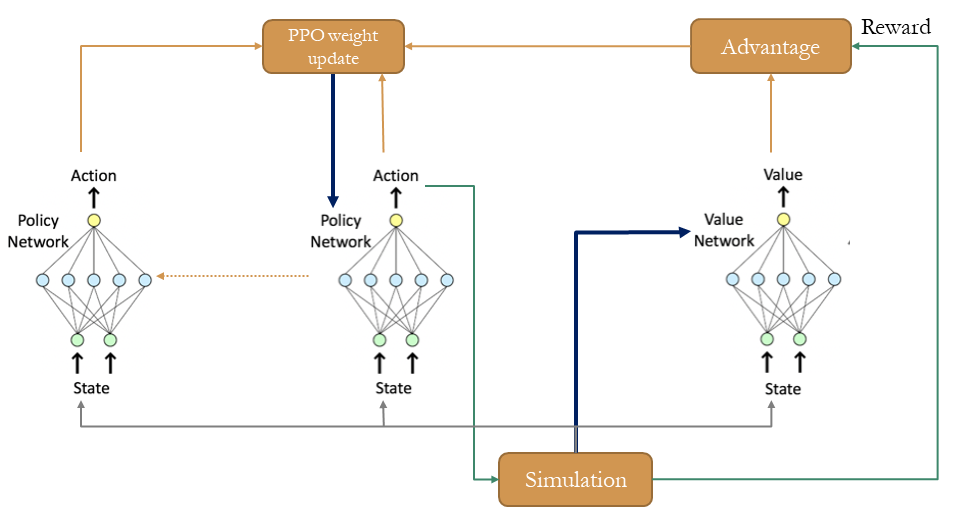}
\caption{An illustration of proximal policy optimiser}
\label{fig:ppo}
\end{figure}

The processing of policy optimisation is shown in Fig. \ref{fig:ppo}. The PPO algorithm in A2C style is explained in detail in \cite{schulman2017proximal}. 

\textbf{Combining Intrinsic and Extrinsic Reward:}
For this problem of taxi routing, the extrinsic reward is defined by bestowing a unit reward for every time step when a commuter is in taxi, for each taxi operating in the road network, characterising the optimisation objective defined in Sec.\ref{sec:problem}. There is no punishment for vacant taxis cruising. The extrinsic reward at each time $t$ is formally defined as cumulative future reward from time $t$ to finite horizon $H$, shown in the following Eq.(\ref{eq:e_reward}). 

\begin{align}
   & R_e(s_t, a_t)  = \sum_t^{H}\sum_{v \in V}(o_v) \\
    & o_v  = \begin{cases}
        1, \text{when vehicle $v$ is occupied}\\
        0, \text{when vehicle $v$ is vacant}\\
    \end{cases}
\label{eq:e_reward} 
\end{align}


Well-performed policy in RL maintain a right balance of maximising extrinsic rewards (exploitation) and trying out new states (exploration), especially when the extrinsic rewards are sparse and the target task is temporally extended. Without proper exploration method, the proposed RL algorithm may converge to sub-optimal solutions without attempting enough novel actions for potentially higher expected rewards. Though finding optimal exploration strategies under large, infinite MDPs (like this problem) is theoretically intractable, many exploration strategies are proposed to practically solve this problem, including optimistic exploration \cite{burda2018exploration, fu2017ex2, tang2017exploration}, Thompson sampling style algorithms \cite{osband2016deep}, information gain style algorithms \cite{houthooft2016vime}, etc. We adopt a random network distillation (RND) approach \cite{burda2018exploration} to construct our intrinsic rewards, which is a heuristic estimation of state counts via prediction errors as exploration bonus. Generally, the intuition behind this approach is to encourage the agent to visit novel states by telling if the current state is novel or not. The RND approach takes inspirations of count-based optimistic exploration and information gain approach. This approach can also handle large input states where most states are visited at most once, as compared to the other approaches.

Specifically, the RND includes two networks: a target network and a predictor network. The target network takes input state to an embedding $f^*(s, a)=f_\psi(s,a)$, where $\psi$ is a random parameter vector. The predictor network is trained by gradient descent to minimise the Mean Squared Error (MSE) between output prediction $\hat f_\phi (s,a)$ and the embedding $f^\psi(s,a)$. The prediction error $R_i (s,a) = ||\hat f_\phi (s,a) - f_\psi (s,a)||^2$ is set as the intrinsic reward (exploration bonus). 

The final reward function is a linear combination of the extrinsic reward and intrinsic reward. The reward function proposed does not tell the vehicles anything about where the commuters are and how to reach the potential commuters. Further normalisation has been done for the intrinsic reward to divide it by a running estimate of standard deviation, so as to keep the intrinsic reward within a reasonable scale when combined with the extrinsic reward. The mathematical representation of the combined reward is in Eq.(\ref{eq:reward}), where $R_e$ and $R_i$ are extrinsic reward and intrinsic reward respectively, $n_s$ represents the number of total observed states and the corresponding number of actions, $\sigma$ is the normalisation parameter, and $
R$ is the total reward.

\begin{align}
    & \sigma  = \sqrt{\frac{n_s\sum_{j=0}^{n_s} R_{i}(s_j, a_j)^2 - (\sum_{j=0}^{n_s} R_{i}(s_j, a_j))^2}{n_s(n_s-1)}} \\
    & R(s_t, a_t)  = R_e(s_t, a_t) + \frac{R_i(s_t, a_t)}{\sigma}
    \label{eq:reward}
\end{align}

\subsubsection{Initialising Model-free Reinforcement Learning}
In practice, the model-based routing policies leverage the full knowledge of the environment and the MDP, but the best-performing dispatch algorithm (non-committed normalised dispatch algorithm as shown in Sec.\ref{sec:mb result}) still in most cases lags behind the pure model-free learners especially when the road network size is small and/or the normalised entropy of given demand pattern is also relatively small. The reasons may lie in the mismatch between the optimal state-action distribution and the dispatch algorithm guided state-action distribution as a result of open-loop control algorithm. 

To combine the benefits of model-free learners and model-based routing policies, we first use imitation learning to train a model-free RL agent to imitate the model-based routing policy as an initialisation, and then further train the model-free learner to gather on-policy data in order to minimise the mismatch between the gathered data state-action distribution and the learned state-action distribution. The model-based routing policy $\mathcal{P}_{ij}$ generated from Sec.\ref{sec: MB routing} is first collected to train a initial policy of our model-free learner described in Sec.\ref{sec: MF RL}. The policy network parameterised by $\theta$ is trained to match the trained policy $\hat{\mathcal P}_{ij}\left(\theta \right)$ to the target effective policy $\mathcal{P}_{ij}$. Let $n$ be the number of nodes in road network $G$, and let $J$ represent the number of direct connections from $i$ where $i \neq j$ and $\mathcal A_{ij} \neq 0$, the objective is to minimise cross entropy loss as Eq.(\ref{eq:ce}) between the model-based target policy and trained policy, using the Adam optimiser.

\begin{equation}
    \operatornamewithlimits{argmin}_\theta \left( -\sum_{i=0}^{n} \sum_{j=0}^{J}\mathcal{P}_{ij} \log{\hat{\mathcal P}_{ij}\left(\theta \right)}\right)
    \label{eq:ce}
\end{equation}

The trained policy is used as initial policy of our model-free learner. After initialisation, on-policy data is collected at each optimisation iteration to update learned policy $\mathcal{P}_{ij} \left(\theta\right)$ parameterised by $\theta$ using model-free RL proposed. The process is carried out until the optimisation objective based on average hiring time of taxis for each second meets expectation.

\section{Experiments}
\label{sec:experiment}

The performance of our proposed reinforcement learning methods have been evaluated on large scale simulation platform described in Sec.\ref{sec:simulation}. Linear function approximators, optimised by the Adam optimiser, are used for both value network and policy network. 
At each epoch, a large number of instances consisting of vacant taxis' positions, choices of actions, as well as the extrinsic rewards of such actions over horizon $H$, are generated from the simulation.

\begin{figure}[ht]
\centering
\includegraphics[width=9cm]{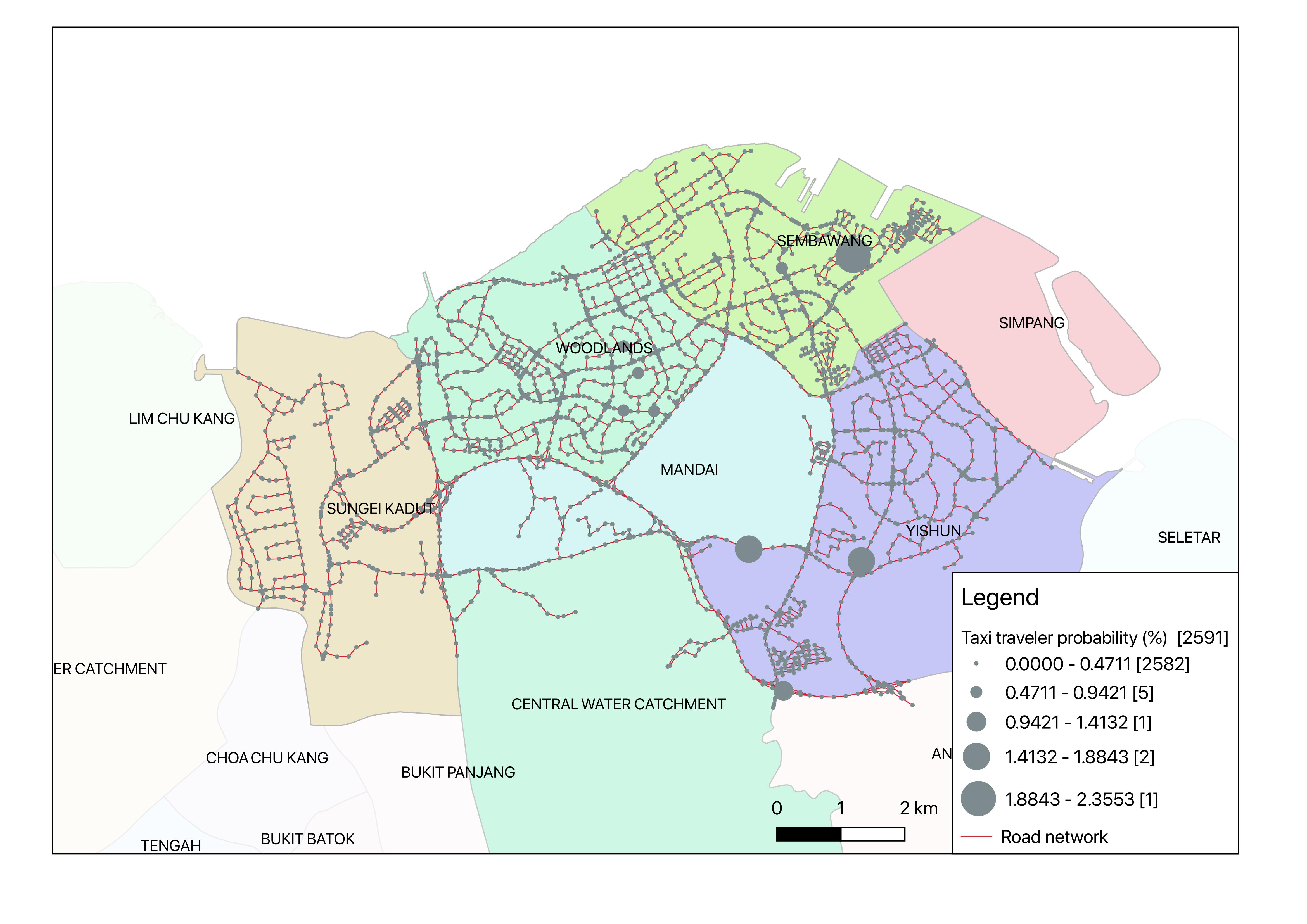}
\caption{Community-based Singapore road network for experiment. The size of the node represents $g_i$, indicating the probability of generating a commuter at each second in percentage.}
\label{fig:network}
\end{figure}

Experiments were conducted on two types of datasets: an artificial road network where the commuter demands $g_i$ and origins/destinations $M_{ij}$ were randomly generated, and a community-based Singapore road network with both random and empirical commuter demands $g_i$. The community-based Singapore road network covers the entire Woodlands, Sembawang, Simpang, Mandai and Yishun areas at the north of Singapore, as shown in Figure \ref{fig:network}. The network consists of $2591$ nodes and $5833$ edges. Empirical commuter demands $g_i$ were generated from trip data of a weekday in $2018$.

The benchmark tasks include evaluations of the produced routing policy on road network specified in the following Table \ref{tab:benchmark}. The artificial road networks consist of square lattices with random bi-directional edge weights. $g_i$ is non-uniform and generated randomly with normalised entropy $\eta$ within $\left(0.6, 0.8\right)$ representing randomness of the information, calculated by Eq.(\ref{eq:entropy}) and (\ref{eq:entropy2}), where $p_i$ is normalised $g_i$ that sum up to $1$ and $n$ represents the number of nodes of given road network. $M_{ij}$ is generated uniformly --- at each origin $i$ in network, the probability of generating a commuter to any other destination node $j$ is draw from a uniform distribution $unif \sim \{0, 1\}$, and $M_{ii}=0$. Note that with many different realisations of the commuter generation and $M_{ij}$ patterns are tested, the results are qualitatively very similar.

\begin{equation}
    p_i = \frac{g_i}{\sum_{i=1}^{n}g_i}
\label{eq:entropy} 
\end{equation}

\begin{equation}
     \eta \left(g_i\right) = -\frac{\sum_i^n p_i \log p_i}{\log n}
\label{eq:entropy2}
\end{equation}

\begin{table}[ht]
\centering
 \begin{tabular}{||c c c c c c c||} 
 \hline
 Setting & Network Type & Nodes & Taxis& $g_i$ & $\eta\left(g_i\right)$ & $M_{ij}$ \\ [0.5ex] 
 \hline\hline
 S1 & Lattice & 100 & [0, 200]& Random & $0.6649$ &Random \\
 \hline
 S2 & Lattice & 1089 & [0, 200] &Random & $0.7124$ &Random \\
 \hline
 S3 & Road Network & 2591 & [30, 200] &Random &$0.7438$ & Random \\
 \hline
 S4 & Road Network & 2591 & [30, 200] & Empirical & $0.7821$ & Random \\[1ex] 
 \hline
\end{tabular}
\caption{Benchmark Task Settings} \label{tab:benchmark} 
\end{table}

In our implementation, the performance metrics used for evaluation are the average extrinsic reward over the simulation horizon $H$ denoted as Eq.(\ref{eq:avgre}), and average waiting time of all commuters given by Eq.(\ref{eq:avg_waiting}). In general maximising the average reward of taxis will result in minimising the average waiting time of commuters, especially in our experimental scenarios where the origins and the length of the trip have no correlations. For simplification, the average extrinsic reward $\bar{R_e}$ is named as \emph{reward} during the discussion of the results in Sec.\ref{sec:results}.

\begin{equation}
    \bar{R_e} = \frac{R_e(s_0, a_0)}{H}
    \label{eq:avgre}
\end{equation}

The implementations, experimental details and hyper-parameter values are further listed in the Appendix.

\section{Results}
\label{sec:results}

We first examine the design decisions for our model-based approaches and model-free RL, presented in Sec.\ref{sec:mb result} and \ref{sec:mf result}, using artificial road network and randomised $g_i$ and $M_{ij}$. 
We then compared our model-based and model-free approaches, as well as our model-free RL with model-based initialisation to demonstrate the effectiveness of our hybrid approach using both artificial road network and community-based Singapore road network with empirical demands, presented in Sec.\ref{sec:mfmb result}.

\subsection{Model-based Routing Policy from Dispatch Algorithm}
\label{sec:mb result}

\begin{figure}[ht]
\centering
\includegraphics[width=9cm]{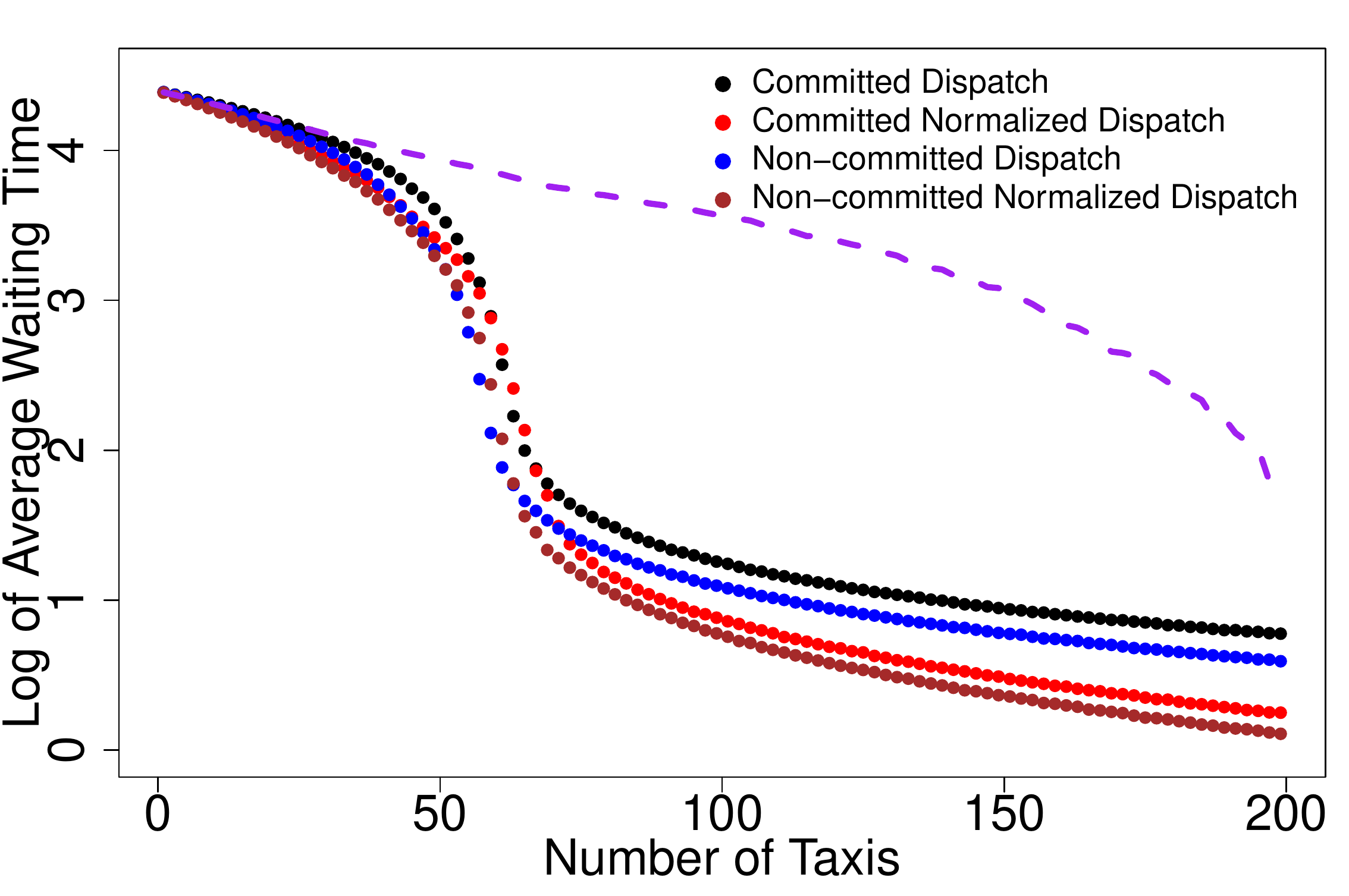}
\caption{Performance comparison of dispatch algorithms on artificial lattice network (experiment setting S1)}
\label{fig:dispatch}
\end{figure}

Multiple dispatch algorithms have been proposed and compared based on 1) whether the taxis are committed to dispatch to the target node to serve commuters (do not serve other commuters during their travels) and 2) whether normalisation factor $\lambda_i$ in Equation \ref{dm1} has been used. Experiments were conducted to evaluate the performance of the proposed dispatch algorithms on artificial network of $100$ nodes (experiment setting S1). The performance of four dispatch algorithms: committed dispatch, committed normalised dispatch, non-committed dispatch, and non-committed normalised dispatch in terms of average waiting time of commuters are shown in Figure \ref{fig:dispatch}. From the comparison, non-committed normalised dispatch algorithm outperforms the other algorithms and is selected to be used to derive effective policy $\mathcal{P}_{ij}$ to initialise model-free RL.

\subsection{Design Decisions for Model-free RL}
\label{sec:mf result}

\begin{figure}[ht]
\centering
\includegraphics[width=12.5cm]{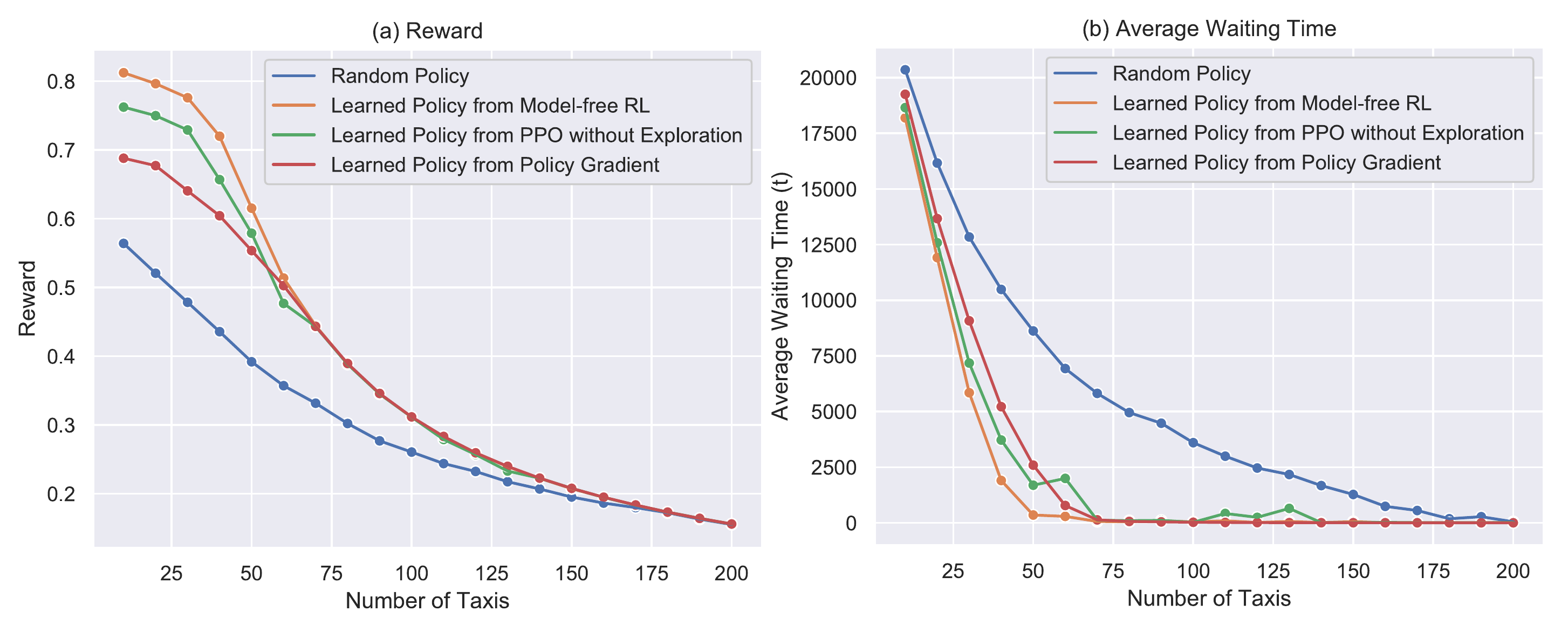} 
\caption{Evaluation of design decisions for our model-free RL on lattice network of $100$ nodes (experiment setting S1). The number of epochs is $1000$. The final reward and average waiting time presented are average results over the last $10$ epochs. (a) final reward (average hiring time of taxis for each second), (b) final average waiting time of commuters.}
\label{fig:comp 100}
\end{figure}

The design decisions for our proposed model-free RL were evaluated by comparing the final reward and average waiting time of commuters from learned policies by vanilla policy gradient, mode-free RL without exploration bonus, and proposed model-free RL algorithm (Figure \ref{fig:comp 100}). All the other design decisions including epochs, learning rate, iterations, clipping range, simulation time, and number of instances collected have been fine-tuned to the best design choices for each method. The results show that our designed model-free algorithm combining intrinsic and extrinsic rewards outperform vanilla policy gradient and the design without exploration bonus, especially when the average waiting time of commuters is affected sub-linearly with increased number of taxis (discussed in detail later). 

\subsection{Evaluation on Artificial Road Network}
\label{sec:mfmb result}

\begin{figure}[bh!]
\centering
\includegraphics[width=12.5cm]{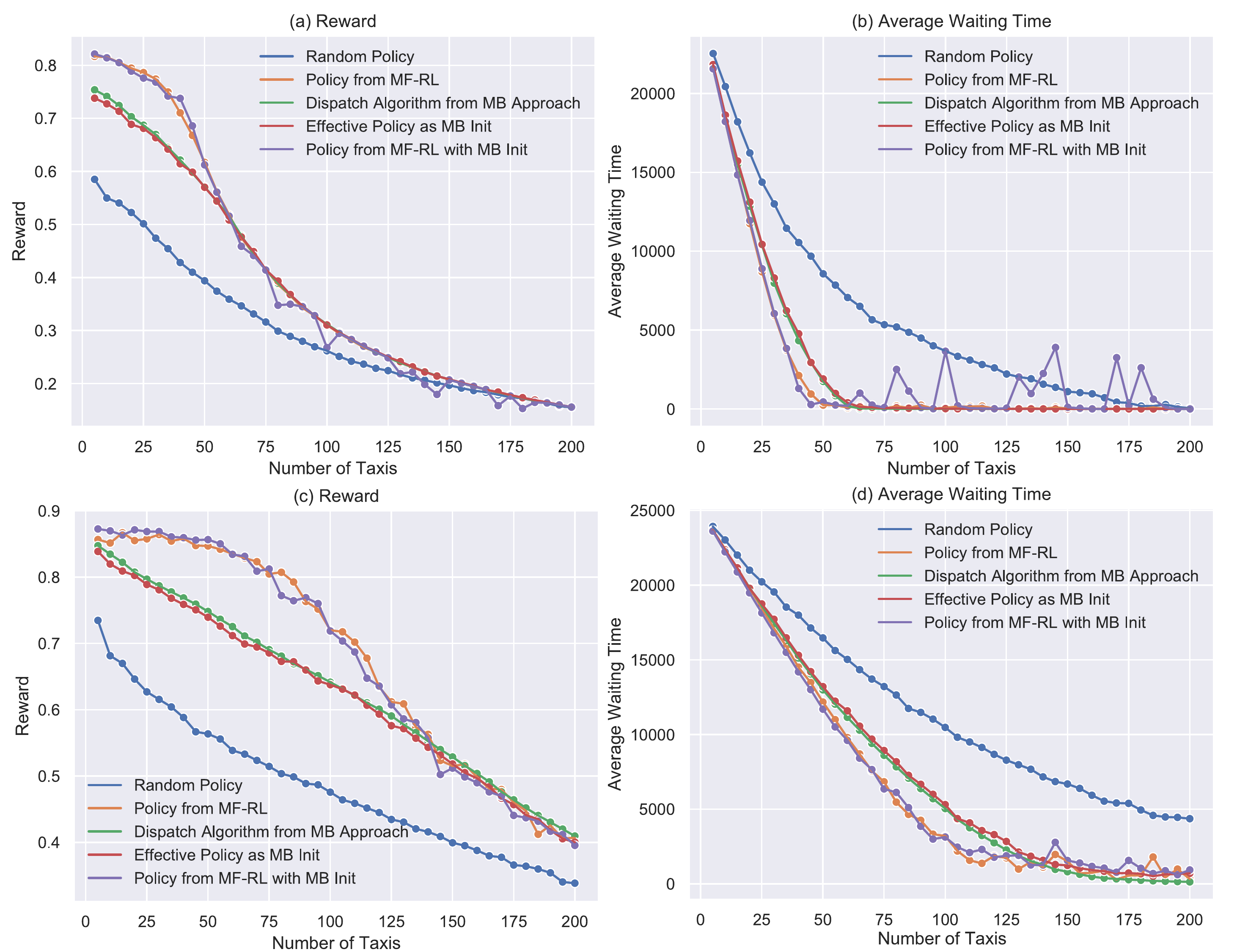} 
\caption{Comparison of performance on artificial road network among random policy, learned policy from model-free RL (MF-RL), dispatch algorithm from model-based approach (MB Approach), effective policy as model-based initialisation (MB Init), and learned policy from model-free RL with model-based initialisation (MF-RL with MB Init). The number of epochs is $1000$. The final reward and average waiting time presented are average results over the last $10$ epochs. (a) and (b) are final reward and average waiting time of commuters experimented on artificial lattice of $100$ nodes (experiment setting S1), (c) and (d) are corresponding results on artificial lattice of $1089$ nodes (experiment setting S2).}
\label{fig:rl_lattice}
\end{figure}

The final performances of our proposed three approaches --- model-based dispatch algorithm, model-free RL and model-free RL with model-based initialisation are evaluated on the same benchmark tasks specified in Sec.\ref{sec:experiment}. In particular, five different policies are compared, including random policy, learned policy from model-free RL, dispatch policy from model-based approach, effective policy for initialisation, and final learned policy from model-free RL with model-based initialisation. The comparison of final results on the artificial road networks (lattice networks of $100$ nodes and $1089$ nodes) with varied number of taxis from $0$ to $200$ is first presented in Figure \ref{fig:rl_lattice}. The results show that the final performance (after training $1000$ epochs) in terms of reward and average waiting time of commuters between learned policies from model-free RL, and model-free RL with initialisation, are similar. Both are better than policy derived from non-committed normalised dispatch algorithm, effective policy and random policy. This is especially true when the average waiting time of commuters are relatively long when demand exceeds supply. 

\begin{figure}[ht]
\centering
\includegraphics[width=12.5cm]{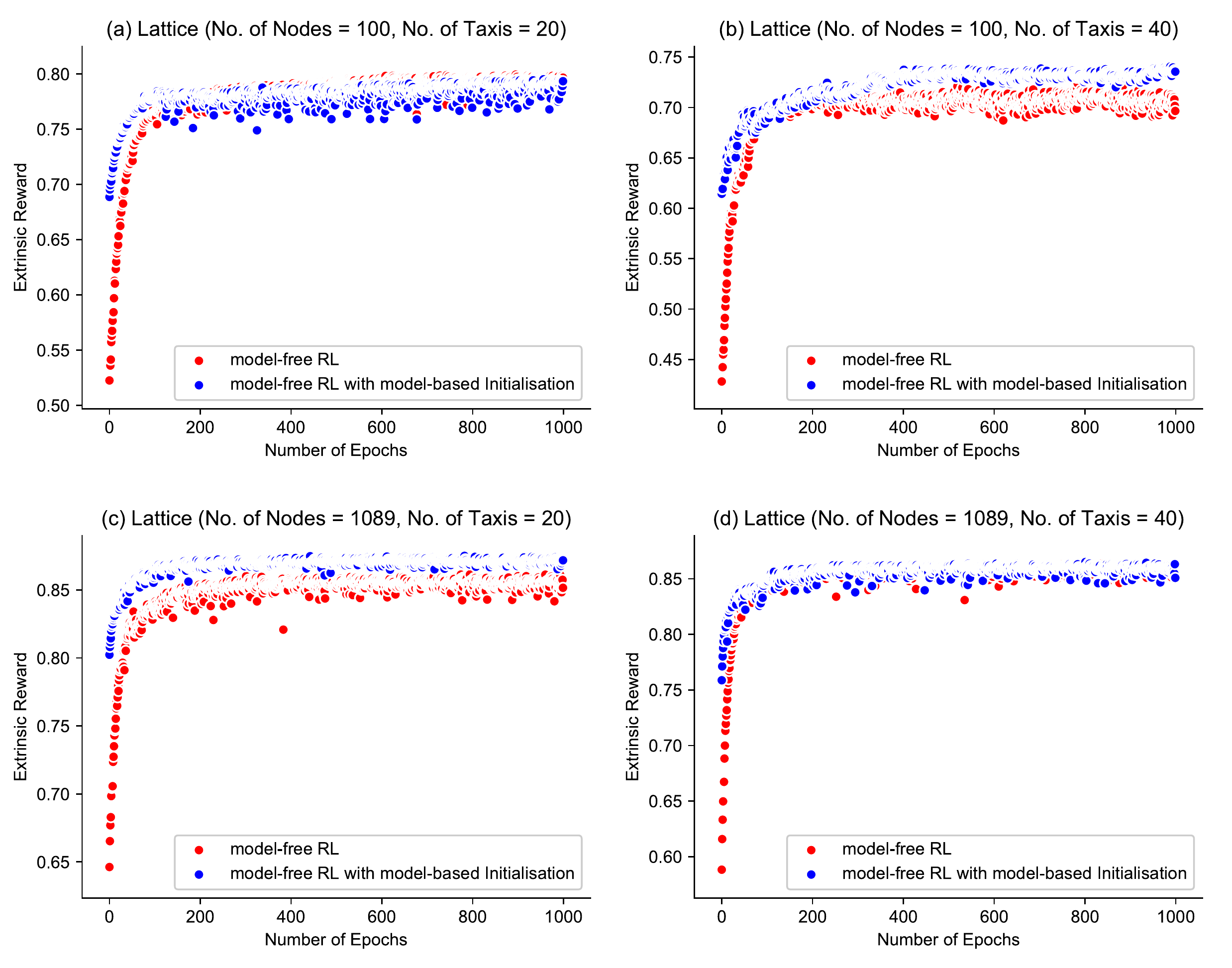} 
\caption{Efficiency evaluation between proposed model-free RL and hybrid approach on artificial road network. (a) and (b) show reward experimented on lattice network of $100$ nodes using $20$ taxis and $40$ taxis respectively, (c) and (d) are corresponding results on lattice network of $1089$ nodes.}
\label{fig:efficiency_lattice}
\end{figure}

Regarding the efficiency of learning, both the model-free RL and the hybrid approach can learn a fairly good policy with better performance than random policy and effective policy from model-based approach very quickly (within $200$ epochs), as shown in Figure \ref{fig:efficiency_lattice}. Even with similar converged performance at the end as in Figure \ref{fig:rl_lattice}, the hybrid approach with model-based initialisation can be much more sample-efficient in learning compared with randomised initialisation. A reasonable reward can be achieved by significantly less training epochs (less than $100$ epochs of training). Note that each epoch entails a simulation run of $50,000$ seconds (simulation time) with maximum $512,000$ data samples collected, detailed in Appendix.  

We have previously identified a transition between the oversaturated phase when demand exceeds supply, and the undersaturated phase when supply exceeds demand, when the structure of road network is given, as discussed in our previous work on phase transition in taxi dynamics \cite{yang2020phase}. The phase transition is also shown in Figure \ref{fig:rl_lattice} that in the oversaturated phase the average waiting time of commuters is affected exponentially with increased number of taxis, while in the undersaturated phase it is affected sub-linearly. In practice, we are more interested in optimising vacant taxi routing in the oversaturated phase to meet the service requirements of commuters efficiently. When the random policy for vacant taxis can already achieve low average waiting time in the undersaturated phase, there is no much room left for optimisation and the redundancy of taxis on roads creates more problems than they save. Specifically for these benchmark tasks, we are mainly evaluating the performance of our proposed algorithm when the number of taxi is less than $50$ for lattice network of $100$ nodes (experiment setting S1) and less than $150$ for lattice network of $1089$ nodes (experiment setting S2).

\subsection{Evaluation on Singapore Road Network}

\begin{figure}[ht]
\centering
\includegraphics[width=12.5cm]{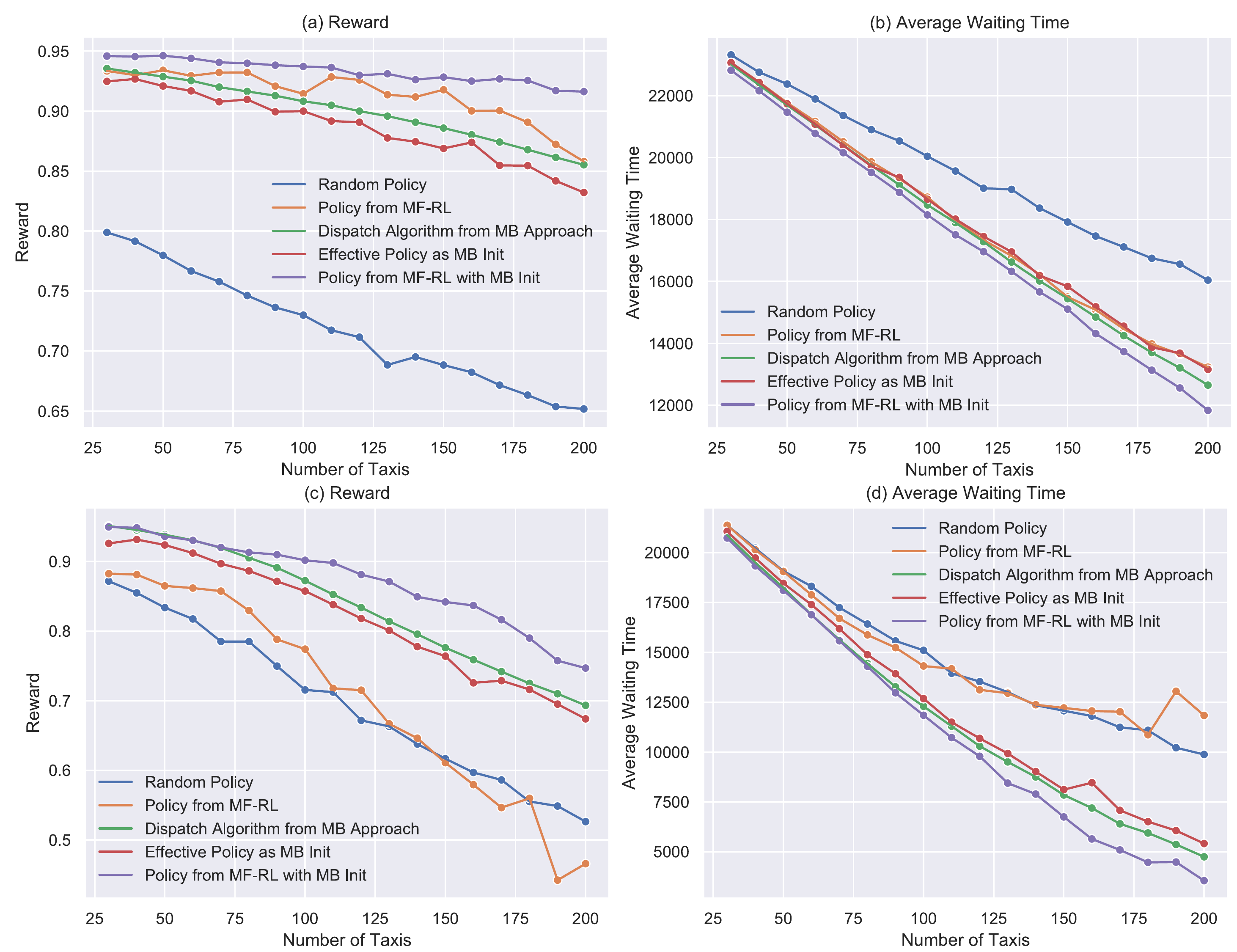} 
\caption{Comparison of performance on community-based Singapore road network among random policy, learned policy from model-free RL (MF-RL), dispatch algorithm from model-based approach (MB Approach), effective policy as model-based initialisation (MB Init) for model-free RL, and learned policy from model-free RL with model-based initialisation (MF-RL with MB Init). The number of epochs is $3000$. The final reward and average waiting time presented are average results over the last $10$ epochs. (a) and (b) are final reward and average waiting time of commuters experimented on randomised demands (experiment setting S3), (c) and (d) are corresponding results on empirical demands (experiment setting S4).}
\label{fig:rl_real}
\end{figure}

The performance of our proposed approaches is also evaluated on real community-based Singapore road network as described in Sec.\ref{sec:experiment}. We compare the results using both randomised demands and empirical demands with taxi numbers ranging from $30$ to $200$, as shown in Figure \ref{fig:rl_real}. Results demonstrate a consistently better performance of proposed hybrid approach, in compare with all the other approaches on a medium-sized real-world road network of $2591$ nodes. Figure \ref{fig:rl_real} also shows that when the normalised entropy of the empirical demands is larger as compared with random demands (the empirical demands of these areas are more uniformly distributed spatially with less ``hotspots''), the number of taxis needed to operate on the road network can be significantly less, but still with higher average hiring time of taxi drivers (reward) and lower average waiting time of commuters when using efficient routing policies proposed in this work.

\begin{figure}[ht]
\centering
\includegraphics[width=12.5cm]{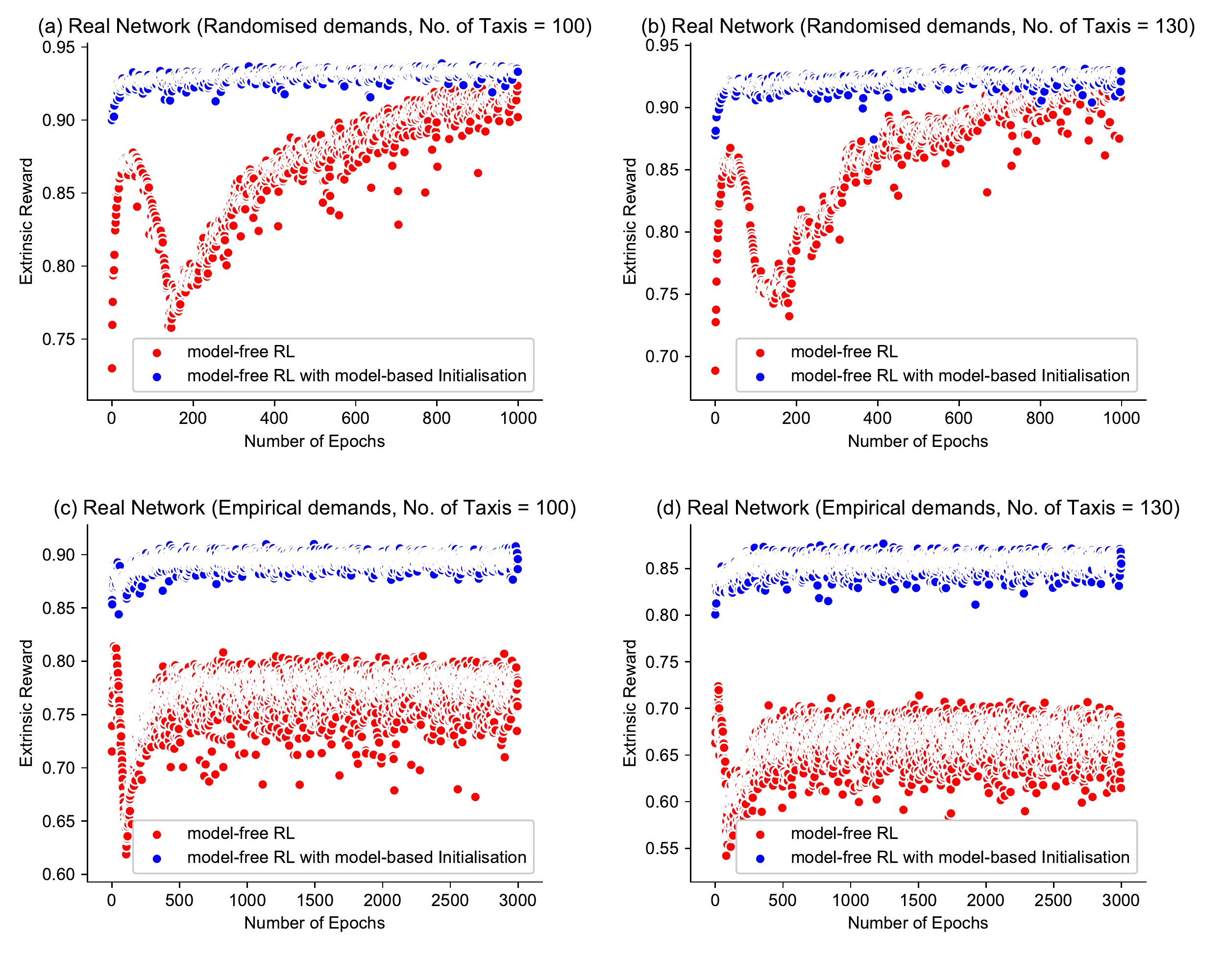} 
\caption{Efficiency evaluation between proposed model-free RL and hybrid approach on community-based Singapore road network. (a) and (b) show reward experimented on same network with random demands using $100$ taxis and $130$ taxis respectively, (c) and (d) are corresponding results with empirical demands.}
\label{fig:efficiency_real}
\end{figure}

By comparing the efficiency of proposed hybrid approach and pure model-free approach on Singapore road network (shown in Figure \ref{fig:efficiency_real}), the hybrid approach can achieve much higher final performance and uses significantly less data samples and training epochs, especially when using empirical demand data (for which the normalised entropy is larger compared to the randomised demands). 

\section{Summary and Outlook}
\label{sec:discussion}

In summary, we presented in this paper methodologies in optimising vacant taxi routing for street-hailing in response to a stochastic demand pattern, so as to maximise the average time of hiring of taxis, and thus reducing the average waiting time of commuters. A model-based dispatch algorithm, a model-free RL based on PPO and RND, and a hybrid RL approach with model-based initialisation are proposed and evaluated under various experimental scenarios using a large-scale agent-based simulation. In general, the proposed model-free RL with model-based initialisation achieves best final performance on both artificial road network and Singapore road network with empirical demands, and are much more sample efficient than pure model-free RL approach. The sample efficient hybrid approach developed here is compelling, especially for useful applications in managing large taxi fleet with human drivers, as well as intelligent routing software for driverless taxis in the near future. The methodologies proposed can be potentially applied to a wide context in urban transportation and logistics problems with stochastic demands, such as dynamic reallocation of shared resources and last-mile delivery. 

Our proposed model-free RL uses only the positions and actions of all taxis with extrinsic rewards to learn and optimise routing policy on a given road network. With training epochs less than $200$, even if the street-hailing demand pattern is not given, our model-free RL can in most cases achieve high performance compared to model policies, including dispatch policy with given demand patterns. This method is good for vacant vehicle routing when the demand patterns and the dynamics of the environment are too complex to be modelled or to be predicted.

The model-based methods we proposed are not as effective as model-free RL on their owns in most of our experimental settings. However, they offer extremely sample efficient way to derive routing policies with reasonably good performance, when the dynamics of the demand-and-supply can be modelled. They are also useful for real-world tasks when real-time optimisation is needed, long-term training is infeasible or data required for model-free RL is hard to get.

Based on the model-free RL and the model-based top-down approach, we then design a hybrid approach using model-free RL with model-based initialisation to combine the benefits of RL and proposed dispatch algorithm. While lattice network is a regular graph, real-world road network with empirical demands may encode more complex dynamics. The proposed hybrid approach is versatile in multiple scenarios under artificial and real road networks of different complexities, and with different size of taxi fleet, which demonstrates its potential in a general setting. The hybrid approach is simple and effective in terms of design and implementation, thus can be widely applied to real-world applications to enhance performance of existing solutions based on heuristics or meta-heuristics, or to accelerate learning process for model-free learner with better performance and less data samples. The on-policy RL approaches proposed in this paper can also be readily extended to consider changing traffic conditions (i.e. congestion) or demand patterns (i.e. peak or off-peak demand) in many real-world problems, by modifying the road network adjacency matrix according to the empirical data and update routing policies accordingly. 




The methodologies presented in this paper could have wider applications beyond taxi systems, for a wide range of empirical and synthetic systems. For future work, we will work on incorporating time-dependent traffic conditions and demand patterns for real-world impact. We will also extend the agent-based simulation and the proposed model-based and model-free approaches to cases with ride-sharing, for the optimal routing and route-matching between shared commuters. Our ultimate goal is to optimise routing of demand-responsive transit system in a hybrid form consisting of real-time booking, street-hailing, as well as ride-sharing.

\section*{Acknowledgements}
This research is supported by the National Research Foundation, Singapore, and the Land Transport Authority under its Urban Mobility Grand Challenge Programme (Award No. UMGC-L005) and the National Research Foundation, Singapore under the NRF fellowship award (NRF-NRFF12-2020-005).
Any opinions, findings and conclusions or recommendations expressed in this material are those of the authors and do not reflect the views of National Research Foundation, Singapore and the Land Transport Authority.


\bibliographystyle{abbrv}
\bibliography{References}

\begin{thebibliography}{10}

\bibitem{agarwal2019optimality}
A.~Agarwal, S.~M. Kakade, J.~D. Lee, and G.~Mahajan.
\newblock Optimality and approximation with policy gradient methods in markov
  decision processes.
\newblock {\em arXiv preprint arXiv:1908.00261}, 2019.

\bibitem{berbeglia2012hybrid}
G.~Berbeglia, J.-F. Cordeau, and G.~Laporte.
\newblock A hybrid tabu search and constraint programming algorithm for the
  dynamic dial-a-ride problem.
\newblock {\em INFORMS Journal on Computing}, 24(3):343--355, 2012.

\bibitem{bongiovanni2019electric}
C.~Bongiovanni, M.~Kaspi, and N.~Geroliminis.
\newblock The electric autonomous dial-a-ride problem.
\newblock {\em Transportation Research Part B: Methodological}, 122:436--456,
  2019.

\bibitem{boutilier1996planning}
C.~Boutilier.
\newblock Planning, learning and coordination in multiagent decision processes.
\newblock In {\em Proceedings of the 6th conference on Theoretical aspects of
  rationality and knowledge}, pages 195--210. Morgan Kaufmann Publishers Inc.,
  1996.

\bibitem{burda2018exploration}
Y.~Burda, H.~Edwards, A.~Storkey, and O.~Klimov.
\newblock Exploration by random network distillation.
\newblock {\em arXiv preprint arXiv:1810.12894}, 2018.

\bibitem{camacho2013model}
E.~F. Camacho and C.~B. Alba.
\newblock {\em Model predictive control}.
\newblock Springer Science \& Business Media, 2013.

\bibitem{coester2018online}
C.~Coester and E.~Koutsoupias.
\newblock The online $ k $-taxi problem.
\newblock {\em arXiv preprint arXiv:1807.06645}, 2018.

\bibitem{colorni2001modeling}
A.~Colorni and G.~Righini.
\newblock Modeling and optimizing dynamic dial-a-ride problems.
\newblock {\em International transactions in operational research},
  8(2):155--166, 2001.

\bibitem{cordeau2007dial}
J.-F. Cordeau and G.~Laporte.
\newblock The dial-a-ride problem: models and algorithms.
\newblock {\em Annals of operations research}, 153(1):29--46, 2007.

\bibitem{coslovich2006two}
L.~Coslovich, R.~Pesenti, and W.~Ukovich.
\newblock A two-phase insertion technique of unexpected customers for a dynamic
  dial-a-ride problem.
\newblock {\em European Journal of Operational Research}, 175(3):1605--1615,
  2006.

\bibitem{coutinhoimpacts}
F.~M. Coutinho and N.~van Oort.
\newblock Impacts of replacing a fixed transit line by a demand responsive
  transit system.

\bibitem{dantzig1959truck}
G.~B. Dantzig and J.~H. Ramser.
\newblock The truck dispatching problem.
\newblock {\em Management science}, 6(1):80--91, 1959.

\bibitem{dehghani2017stochastic}
S.~Dehghani, S.~Ehsani, M.~Hajiaghayi, V.~Liaghat, and S.~Seddighin.
\newblock Stochastic k-server: How should uber work?
\newblock {\em arXiv preprint arXiv:1705.05755}, 2017.

\bibitem{duan2020centralized}
L.~Duan, Y.~Wei, J.~Zhang, and Y.~Xia.
\newblock Centralized and decentralized autonomous dispatching strategy for
  dynamic autonomous taxi operation in hybrid request mode.
\newblock {\em Transportation Research Part C: Emerging Technologies},
  111:397--420, 2020.

\bibitem{feng2017we}
G.~Feng, G.~Kong, and Z.~Wang.
\newblock We are on the way: Analysis of on-demand ride-hailing systems.
\newblock {\em Available at SSRN 2960991}, 2017.

\bibitem{ealing2019}
T.~for London Consultation~Hub.
\newblock Demand responsive bus trial ealing consultation report, 2019.

\bibitem{fu2017ex2}
J.~Fu, J.~Co-Reyes, and S.~Levine.
\newblock Ex2: Exploration with exemplar models for deep reinforcement
  learning.
\newblock In {\em Advances in neural information processing systems}, pages
  2577--2587, 2017.

\bibitem{gschwind2019adaptive}
T.~Gschwind and M.~Drexl.
\newblock Adaptive large neighborhood search with a constant-time feasibility
  test for the dial-a-ride problem.
\newblock {\em Transportation Science}, 53(2):480--491, 2019.

\bibitem{han2016routing}
M.~Han, P.~Senellart, S.~Bressan, and H.~Wu.
\newblock Routing an autonomous taxi with reinforcement learning.
\newblock In {\em Proceedings of the 25th ACM International on Conference on
  Information and Knowledge Management}, pages 2421--2424, 2016.

\bibitem{ho2018survey}
S.~C. Ho, W.~Szeto, Y.-H. Kuo, J.~M. Leung, M.~Petering, and T.~W. Tou.
\newblock A survey of dial-a-ride problems: Literature review and recent
  developments.
\newblock {\em Transportation Research Part B: Methodological}, 111:395--421,
  2018.

\bibitem{hoque2012analysis}
M.~A. Hoque, X.~Hong, and B.~Dixon.
\newblock Analysis of mobility patterns for urban taxi cabs.
\newblock In {\em 2012 international conference on computing, networking and
  communications (ICNC)}, pages 756--760. IEEE, 2012.

\bibitem{houthooft2016vime}
R.~Houthooft, X.~Chen, Y.~Duan, J.~Schulman, F.~De~Turck, and P.~Abbeel.
\newblock Vime: Variational information maximizing exploration.
\newblock In {\em Advances in Neural Information Processing Systems}, pages
  1109--1117, 2016.

\bibitem{jin2014enhancing}
J.~G. Jin, L.~C. Tang, L.~Sun, and D.-H. Lee.
\newblock Enhancing metro network resilience via localized integration with bus
  services.
\newblock {\em Transportation Research Part E: Logistics and Transportation
  Review}, 63:17--30, 2014.

\bibitem{jin2019mobility}
Z.~R. Jin and A.~Z. Qiu.
\newblock Mobility-as-a-service (maas) testbed as an integrated approach for
  new mobility-a living lab case study in singapore.
\newblock In {\em International Conference on Human-Computer Interaction},
  pages 441--458. Springer, 2019.

\bibitem{jittrapirom2019dutch}
P.~Jittrapirom, W.~van Neerven, K.~Martens, D.~Trampe, and H.~Meurs.
\newblock The dutch elderly's preferences toward a smart demand-responsive
  transport service.
\newblock {\em Research in Transportation Business \& Management}, 30:100383,
  2019.

\bibitem{levine2013guided}
S.~Levine and V.~Koltun.
\newblock Guided policy search.
\newblock In {\em International Conference on Machine Learning}, pages 1--9,
  2013.

\bibitem{liang2020automated}
X.~Liang, G.~H. de~Almeida~Correia, K.~An, and B.~van Arem.
\newblock Automated taxis’ dial-a-ride problem with ride-sharing considering
  congestion-based dynamic travel times.
\newblock {\em Transportation Research Part C: Emerging Technologies},
  112:260--281, 2020.

\bibitem{lim2017pickup}
A.~Lim, Z.~Zhang, and H.~Qin.
\newblock Pickup and delivery service with manpower planning in hong kong
  public hospitals.
\newblock {\em Transportation Science}, 51(2):688--705, 2017.

\bibitem{madsen1995heuristic}
O.~B. Madsen, H.~F. Ravn, and J.~M. Rygaard.
\newblock A heuristic algorithm for a dial-a-ride problem with time windows,
  multiple capacities, and multiple objectives.
\newblock {\em Annals of operations Research}, 60(1):193--208, 1995.

\bibitem{miller2017demand}
J.~Miller and J.~P. How.
\newblock Demand estimation and chance-constrained fleet management for ride
  hailing.
\newblock In {\em 2017 IEEE/RSJ International Conference on Intelligent Robots
  and Systems (IROS)}, pages 4481--4488. IEEE, 2017.

\bibitem{mnih2016asynchronous}
V.~Mnih, A.~P. Badia, M.~Mirza, A.~Graves, T.~Lillicrap, T.~Harley, D.~Silver,
  and K.~Kavukcuoglu.
\newblock Asynchronous methods for deep reinforcement learning.
\newblock In {\em International conference on machine learning}, pages
  1928--1937, 2016.

\bibitem{mnih2013playing}
V.~Mnih, K.~Kavukcuoglu, D.~Silver, A.~Graves, I.~Antonoglou, D.~Wierstra, and
  M.~Riedmiller.
\newblock Playing atari with deep reinforcement learning.
\newblock {\em arXiv preprint arXiv:1312.5602}, 2013.

\bibitem{mnih2015human}
V.~Mnih, K.~Kavukcuoglu, D.~Silver, A.~A. Rusu, J.~Veness, M.~G. Bellemare,
  A.~Graves, M.~Riedmiller, A.~K. Fidjeland, G.~Ostrovski, et~al.
\newblock Human-level control through deep reinforcement learning.
\newblock {\em Nature}, 518(7540):529--533, 2015.

\bibitem{munoz2015methodology}
D.~Mu{\~n}oz-Carpintero, D.~S{\'a}ez, C.~E. Cort{\'e}s, and A.~N{\'u}{\~n}ez.
\newblock A methodology based on evolutionary algorithms to solve a dynamic
  pickup and delivery problem under a hybrid predictive control approach.
\newblock {\em Transportation Science}, 49(2):239--253, 2015.

\bibitem{narayanan2020shared}
S.~Narayanan, E.~Chaniotakis, and C.~Antoniou.
\newblock Shared autonomous vehicle services: A comprehensive review.
\newblock {\em Transportation Research Part C: Emerging Technologies},
  111:255--293, 2020.

\bibitem{nunez2014multiobjective}
A.~N{\'u}{\~n}ez, C.~E. Cort{\'e}s, D.~S{\'a}ez, B.~De~Schutter, and
  M.~Gendreau.
\newblock Multiobjective model predictive control for dynamic pickup and
  delivery problems.
\newblock {\em Control Engineering Practice}, 32:73--86, 2014.

\bibitem{osband2016deep}
I.~Osband, C.~Blundell, A.~Pritzel, and B.~Van~Roy.
\newblock Deep exploration via bootstrapped dqn.
\newblock In {\em Advances in neural information processing systems}, pages
  4026--4034, 2016.

\bibitem{oxley1980dial}
P.~Oxley.
\newblock Dial/a/ride: a review.
\newblock {\em Transportation Planning and Technology}, 6(3):141--148, 1980.

\bibitem{perera2019resurgence}
S.~Perera, C.~Ho, and D.~Hensher.
\newblock Resurgence of demand responsive transit services--insights from bridj
  trials in inner west of sydney, australia, 2019.

\bibitem{reinhardt2013synchronized}
L.~B. Reinhardt, T.~Clausen, and D.~Pisinger.
\newblock Synchronized dial-a-ride transportation of disabled passengers at
  airports.
\newblock {\em European Journal of Operational Research}, 225(1):106--117,
  2013.

\bibitem{schulman2015trust}
J.~Schulman, S.~Levine, P.~Abbeel, M.~Jordan, and P.~Moritz.
\newblock Trust region policy optimization.
\newblock In {\em International conference on machine learning}, pages
  1889--1897, 2015.

\bibitem{schulman2017proximal}
J.~Schulman, F.~Wolski, P.~Dhariwal, A.~Radford, and O.~Klimov.
\newblock Proximal policy optimization algorithms.
\newblock {\em arXiv preprint arXiv:1707.06347}, 2017.

\bibitem{vsemrov2016reinforcement}
D.~{\v{S}}emrov, R.~Marseti{\v{c}}, M.~{\v{Z}}ura, L.~Todorovski, and A.~Srdic.
\newblock Reinforcement learning approach for train rescheduling on a
  single-track railway.
\newblock {\em Transportation Research Part B: Methodological}, 86:250--267,
  2016.

\bibitem{silver2017mastering}
D.~Silver, J.~Schrittwieser, K.~Simonyan, I.~Antonoglou, A.~Huang, A.~Guez,
  T.~Hubert, L.~Baker, M.~Lai, A.~Bolton, et~al.
\newblock Mastering the game of go without human knowledge.
\newblock {\em Nature}, 550(7676):354--359, 2017.

\bibitem{sutton2018reinforcement}
R.~S. Sutton and A.~G. Barto.
\newblock {\em Reinforcement learning: An introduction}.
\newblock MIT press, 2018.

\bibitem{sutton2000policy}
R.~S. Sutton, D.~A. McAllester, S.~P. Singh, and Y.~Mansour.
\newblock Policy gradient methods for reinforcement learning with function
  approximation.
\newblock In {\em Advances in neural information processing systems}, pages
  1057--1063, 2000.

\bibitem{tang2017exploration}
H.~Tang, R.~Houthooft, D.~Foote, A.~Stooke, O.~X. Chen, Y.~Duan, J.~Schulman,
  F.~DeTurck, and P.~Abbeel.
\newblock \# exploration: A study of count-based exploration for deep
  reinforcement learning.
\newblock In {\em Advances in neural information processing systems}, pages
  2753--2762, 2017.

\bibitem{teodorovic2000fuzzy}
D.~Teodorovic and G.~Radivojevic.
\newblock A fuzzy logic approach to dynamic dial-a-ride problem.
\newblock {\em Fuzzy sets and systems}, 116(1):23--33, 2000.

\bibitem{verma2017augmenting}
T.~Verma, P.~Varakantham, S.~Kraus, and H.~C. Lau.
\newblock Augmenting decisions of taxi drivers through reinforcement learning
  for improving revenues.
\newblock In {\em Twenty-Seventh International Conference on Automated Planning
  and Scheduling}, 2017.

\bibitem{wang2018understanding}
Y.~Wang, B.~Zheng, and E.-P. Lim.
\newblock Understanding the effects of taxi ride-sharing—a case study of
  singapore.
\newblock {\em Computers, Environment and Urban Systems}, 69:124--132, 2018.

\bibitem{yang2018turn}
B.~Yang and Q.~Li.
\newblock Turn-by-turn intelligent manoeuvring of driverless taxis: A recursive
  value model enhanced by reinforcement learning.
\newblock In {\em 2018 IEEE Intelligent Vehicles Symposium (IV)}, pages
  1659--1664. IEEE, 2018.

\bibitem{yang2020phase}
B.~Yang, S.~Ren, E.~F. Legara, Z.~Li, E.~Y. Ong, L.~Lin, and C.~Monterola.
\newblock Phase transition in taxi dynamics and impact of ridesharing.
\newblock {\em Transportation Science}, 2020.

\bibitem{yu2019markov}
X.~Yu, S.~Gao, X.~Hu, and H.~Park.
\newblock A markov decision process approach to vacant taxi routing with
  e-hailing.
\newblock {\em Transportation Research Part B: Methodological}, 121:114--134,
  2019.

\bibitem{zhang2019multi}
K.~Zhang, Z.~Yang, and T.~Ba{\c{s}}ar.
\newblock Multi-agent reinforcement learning: A selective overview of theories
  and algorithms.
\newblock {\em arXiv preprint arXiv:1911.10635}, 2019.

\bibitem{zhu2017target}
Y.~Zhu, R.~Mottaghi, E.~Kolve, J.~J. Lim, A.~Gupta, L.~Fei-Fei, and A.~Farhadi.
\newblock Target-driven visual navigation in indoor scenes using deep
  reinforcement learning.
\newblock In {\em 2017 IEEE international conference on robotics and automation
  (ICRA)}, pages 3357--3364. IEEE, 2017.

\end{thebibliography}

\newpage

\appendix
\section{Implementation Details}

The simulation was designed for infinite time horizons and infinite number of runs. For each run, the unit for the length of simulation is defined as ``Second'' for simplicity. The settings that are consistent for both model-based approaches and model-free approaches are listed below.

\begin{table}[h!]
\centering
 \begin{tabular}{|c | c|} 
 \hline
 Length of each simulation run & $50, 000$ \\
 \hline
 Maximum number of commuters waiting at each node& $1, 000, 001$ \\
 \hline
Maximum sample size collected for each run & $512, 000$ \\
\hline

\end{tabular}
\label{tab:sim_parameters} 
\end{table}

The imitation learning used to initialise model-free leaner will update the policy network until the cross entropy loss stop decreasing or reach maximum update steps $30, 000$. The hyperparameters used in training the model-free RL and the hybrid approach are the same for all experiment settings and are listed below.

\begin{table}[h!]
\centering
 \begin{tabular}{|c | c|} 
 \hline
 Learning rate & $0.01$ \\
 \hline
 Clipping parameter epsilon& $0.1$ \\
 \hline
Iterations & $10$ \\
\hline
Batch size &  Input data size\\
\hline

\end{tabular}
\label{tab:imitation_parameters} 
\end{table}

\section{Experimental Details}

The experimental details for both artificial road network and community-based Singapore road network are listed below. All experiments were conducted on a commodity server with two NVIDIA Quadro P4000 GPUs with $8$ GB GDDR5 memory.

\begin{table}[h!]
\centering
 \begin{tabular}{|c| c| c | c|} 
 \hline
 Experiment & Road Network Type &Edge Travel Time & Epochs \\
 \hline
 S1 & $10 \times 10$ Lattice & $X\sim unif\{1, 10\}$ & $1000$  \\
 \hline
 S2& $33 \times 33$ Lattice& $X\sim unif\{1, 10\}$  & $1000$\\
 \hline
S3 & Road network ($2591$ nodes) & Travel time (s) & $1000$ \\
\hline
S4 & Road network ($2591$ nodes) &Travel time (s)&  $3000$\\
\hline

\end{tabular}
\label{tab:exp_parameters} 
\end{table}

\label{sec:appendix}

\end{document}